%% file: main.tex
\documentclass{article}
\PassOptionsToPackage{numbers, compress}{natbib}
\usepackage[final]{bdl_2018}

\usepackage[utf8]{inputenc}
\usepackage{amsmath,amssymb} \usepackage{bm}
\usepackage{xspace}
\usepackage{booktabs}
\usepackage{subcaption}
\usepackage{array,graphicx}
\usepackage{color}
\usepackage{hyperref}
\usepackage{xcolor}
\hypersetup{
    colorlinks,
    anchorcolor={blue!50!black},
    linkcolor={blue!50!black},
    citecolor={blue!50!black},
    urlcolor={blue}
}
\usepackage{footmisc}
\usepackage{tablefootnote}
\usepackage{enumitem}   
\setlist[enumerate]{itemsep=0pt,topsep=0pt,leftmargin=1cm}

\usepackage[tableposition=top]{caption}
\DeclareCaptionLabelFormat{andtable}{#1~#2  \&  \tablename~\thetable}

\newcommand{\mc}{\mathcal}

\newcommand{\ex}{{\mathbb E}}
\newcommand{\exe}{\ex_{\hat p(x)}}
\newcommand{\exey}{\ex_{\hat p(x,y)}}
\newcommand{\dkl}{D_\text{KL}}
\newcommand{\Lvae}{\mc L_\text{VAE}}
\newcommand{\Lae}{\mc L_\text{AE}}
\newcommand{\tc}{\text{TC}}
\newcommand{\hsic}{\text{HSIC}}
\newcommand{\mmd}{\text{MMD}}
\newcommand{\cov}{\text{Cov}}
\newcommand{\diag}{\text{diag}}
\newcommand{\C}{\mathcal{C}}

\newcommand{\N}{\mc N}
\newcommand*\diff{\mathop{}\!\mathrm{d}}

\usepackage{pgfplots}
\usepackage{tikz}
\usepackage{xcolor}
\usepackage{placeins}

\usepackage{amsthm}

\newcommand\crule[3][black]{\textcolor{#1}{\rule{#2}{#3}}}

\newcommand{\hlm}[1]{\paragraph{#1}}

\author{
  Michael Tschannen \\
  ETH Zurich \\
  \texttt{michaelt@nari.ee.ethz.ch} \\
  \And
  Olivier Bachem \\
  Google AI, Brain Team \\
  \texttt{bachem@google.com} \\
  \And
  Mario Lucic \\
  Google AI, Brain Team \\
  \texttt{lucic@google.com}
}

\title{Recent Advances in Autoencoder-Based Representation Learning}
\begin{document}
\maketitle

\begin{abstract}
Learning useful representations with little or no supervision is a key challenge in artificial intelligence. We provide an in-depth review of recent advances in representation learning with a focus on autoencoder-based models. To organize these results we make use of \emph{meta-priors} believed useful for downstream tasks, such as disentanglement and hierarchical organization of features. In particular, we uncover three main mechanisms to enforce such properties, namely (i) regularizing the (approximate or aggregate) posterior distribution, (ii) factorizing the encoding and decoding distribution, or (iii) introducing a structured prior distribution. While there are some promising results, implicit or explicit supervision remains a key enabler and all current methods use strong inductive biases and modeling assumptions. Finally, we provide an analysis of autoencoder-based representation learning through the lens of rate-distortion theory and identify a clear tradeoff between the amount of prior knowledge available about the downstream tasks, and how useful the representation is for this task.
\end{abstract}

\section{Introduction}
The ability to learn useful representations of data with little or no supervision is a key challenge towards applying artificial intelligence to the vast amounts of unlabelled data collected in the world.
While it is clear that the usefulness of a representation learned on data heavily depends on the end task which it is to be used for, one could imagine that there exists properties of representations which are useful for many real-world tasks simultaneously. In a seminal paper on representation learning \citet{bengio2013representation} proposed such a set of \emph{meta-priors}.
The meta-priors are derived from general assumptions about the world such as the hierarchical organization or disentanglement of explanatory factors, the possibility of semi-supervised learning, the concentration of data on low-dimensional manifolds, clusterability, and temporal and spatial coherence.

Recently, a variety of (unsupervised) representation learning algorithms have been proposed based on the idea of \emph{autoencoding} where the goal is to learn a mapping from high-dimensional observations to a lower-dimensional representation space such that the original observations can be reconstructed (approximately) from the lower-dimensional representation. While these approaches have varying motivations and design choices, we argue that essentially all of the methods reviewed in this paper implicitly or explicitly have at their core at least one of the meta-priors from \citet{bengio2013representation}.

Given the unsupervised nature of the upstream representation learning task, the characteristics of the meta-priors enforced in the representation learning step determine how useful the resulting representation is for the real-world end task. 
Hence, it is critical to understand which meta-priors are targeted by which models and which generic techniques are useful to enforce a given meta-prior. In this paper, we provide a unified view which encompasses the majority of proposed models and relate them to the meta-priors proposed by \citet{bengio2013representation}. We summarize the recent work focusing on the meta-priors in Table~\ref{tab:metaprior}.

\paragraph{Meta-priors of \citet{bengio2013representation}.} Meta-priors capture very general premises about the world and are therefore arguably useful for a broad set of downstream tasks. We briefly summarize the most important meta-priors which are targeted by the reviewed approaches.
\begin{enumerate}
    \item {\bf Disentanglement:} Assuming that the data is generated from independent factors of variation, for example object orientation and lighting conditions in images of objects, disentanglement as a meta-prior encourages these factors to be captured by different independent variables in the representation. It should result in a concise abstract representation of the data useful for a variety of downstream tasks and promises improved sample efficiency.
    \item {\bf Hierarchical organization of explanatory factors:} The intuition behind this meta-prior is that the world can be described as a hierarchy of increasingly abstract concepts. For example natural images can be abstractly described in terms of the objects they show at various levels of granularity. Given the object, a more concrete description can be given by object attributes.
    \item {\bf Semi-supervised learning:} The idea is to share a representation between a supervised and an unsupervised learning task which often leads to synergies: While the number of labeled data points is usually too small to learn a good predictor (and thereby a representation), training jointly with an unsupervised target allows the supervised task to learn a representation that generalizes, but also guides the representation learning process.
    \item {\bf Clustering structure:} Many real-wold data sets have multi-category structure (such as images showing different object categories), with possibly category-dependent factors of variation. Such structure can be captured with a latent mixture model where each mixture component corresponds to one category, and its distribution models the factors of variation within that category. This naturally leads to a representation with clustering structure.
\end{enumerate}
Very generic concepts such as \emph{smoothness} as well as \emph{temporal and spatial coherence} are not specific to unsupervised learning and are used in most practical setups (for example weight decay to encourage smoothness of predictors, and convolutional layers to capture spatial coherence in image data). We discuss the implicit supervision used by most approaches in Section~\ref{sec:critique}.

\begin{table}[t]
    \centering
    \caption{\label{tab:invidx}Grouping of methods according to the meta-priors for representation learning from \cite{bengio2013representation}. While many methods directly or indirectly address multiple meta-priors, we only considered the most prominent target of each method. Note that meta-priors such as low dimensionality and manifold structure are enforced by essentially all methods. \vspace{2mm}}
    \input{tables/metaprior.tex}
   \label{tab:metaprior}
\end{table}

\paragraph{Mechanisms for enforcing meta-priors.} We identify the following three mechanisms to enforce meta-priors:
\begin{enumerate}[label=(\roman*)]
\item Regularization of the encoding distribution (Section~\ref{sec:regbased}).
\item Choice of the encoding and decoding distribution or model family (Section~\ref{sec:structarch}).
\item Choice of a flexible prior distribution of the representation (Section~\ref{sec:structprior}).
\end{enumerate}
For example, regularization of the encoding distribution is often used to encourage disentangled representations. Alternatively, factorizing the encoding and decoding distribution in a hierarchical fashion allows us to impose a hierarchical structure to the representation. Finally, a more flexible prior, say a mixture distribution, can be used to encourage clusterability.

Before starting our overview, in Section~\ref{sec:prelim} we present the main concepts necessary to understand variational autoencoders (VAEs) \cite{kingma2013auto,rezende2014stochastic}, underlying most of the methods considered in this paper, and several techniques used to estimate divergences between probability distributions. We then present a detailed discussion of regularization-based methods in Section~\ref{sec:regbased}, review methods relying on structured encoding and decoding distributions in Section~\ref{sec:structarch}, and present methods using a structured prior distribution in Section~\ref{sec:structprior}. We conclude the review section by an overview of related methods such as cross-domain representation learning \cite{liu2017detach,lee2018diverse,gonzalez2018image} in Section~\ref{sec:otherapp}. Finally, we provide a critique of unsupervised representation learning through the rate-distortion framework of~\citet{alemi2018fixing} and discuss the implications in Section~\ref{sec:critique}.

\begin{figure}[t]
  \centering
  \begin{subfigure}[b]{.6\linewidth}
    \includegraphics[clip, trim=3.2cm 20cm 3.7cm 3cm, width=1\textwidth]{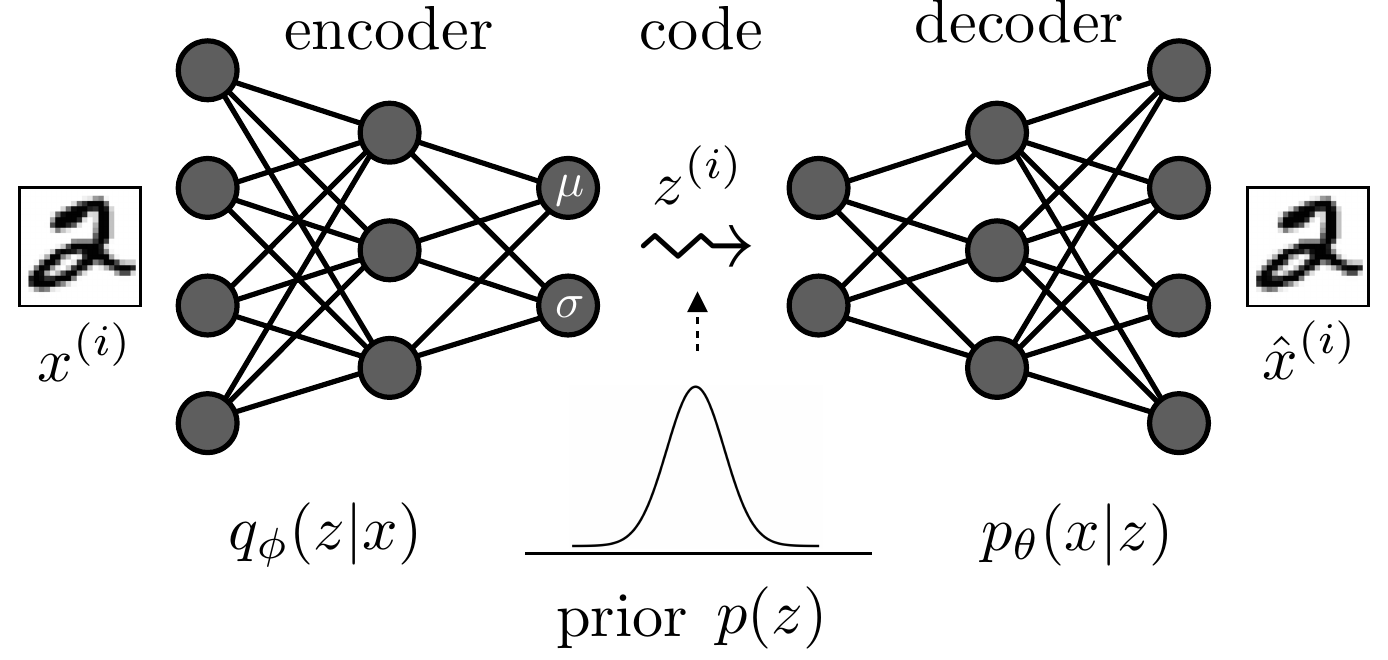}
    \subcaption{Variational Autoencoder (VAE) framework. \label{fig:autoencoder_framework}}
  \end{subfigure}\\[0.3cm]
  \begin{subfigure}[b]{.6\linewidth}
    \includegraphics[width=0.54\textwidth]{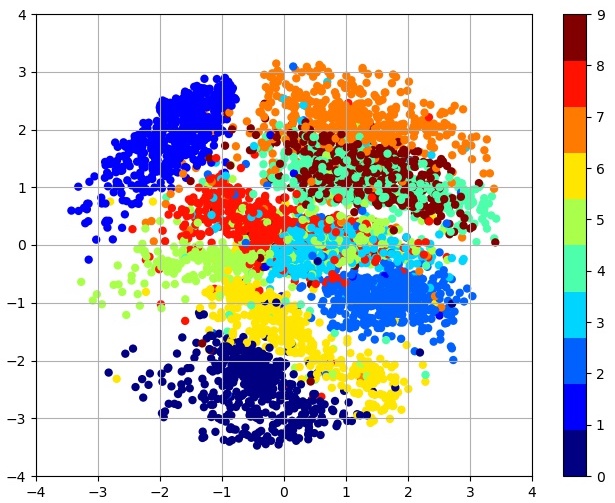}~
    \includegraphics[width=0.44\textwidth]{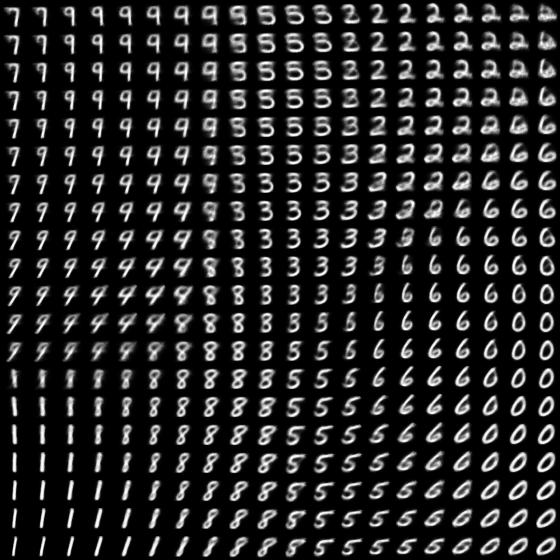}
    \subcaption{Samples from a trained VAE. \label{fig:autoencoder_samples}}
  \end{subfigure}
  \caption{Figure (a) illustrates the Variational Autoencoder (VAE) framework specified by the encoder, decoder, and the prior distribution on the latent (representation/code) space. The encoder maps the input to the representation space (\emph{inference}), while the decoder reconstructs the original input from the representation. The encoder is encouraged to satisfy some structure on the latent space (e.g., it should be disentangled). Figure (b) shows samples from a trained autoencoder with latent space of 2 dimensions on the MNIST data set. Each point on the left corresponds to the representation of a digit (originally in 784 dimensions) and the reconstructed digits can be seen on the right. One can observe that in this case the latent representation is clustered (various styles of the same digit are close w.r.t. $L_2$-distance, and within each group the position corresponds to the rotation of the digit).\label{fig:autoencoder}}
\end{figure}

\section{Preliminaries} \label{sec:prelim}
We assume familiarity with the key concepts in Bayesian data modeling. For a gentle introduction to VAEs we refer the reader to~\cite{doersch2016tutorial}. 
VAEs~\cite{kingma2013auto,rezende2014stochastic} aim to learn a parametric latent variable model 
by maximizing the marginal $\log$-likelihood of the training data $\{x^{(i)}\}_{i=1}^N$. By introducing an approximate posterior $q_\phi(z|x)$ which is an approximation of the intractable true posterior $p_\theta(z|x)$ we can rewrite the negative $\log$-likelihood as
\begin{equation*}
\exe[-\log p_\theta(x)] = \Lvae(\theta, \phi) - \exe[\dkl(q_\phi(z|x)\|p_\theta(z|x))]
\end{equation*}
where
\begin{equation} \label{eq:lvae}
 \Lvae(\theta, \phi) = \exe[ \ex_{q_\phi(z|x)}[- \log p_\theta(x|z)] + \exe[\dkl(q_\phi(z|x) \| p(z))],
\end{equation}
and $\exe[f(x)] = \frac1N \sum_{i=1}^N f(x^{(i)})$ is the expectation of the function $f(x)$ w.r.t. the empirical data distribution. The approach is illustrated in Figure~\ref{fig:autoencoder}.
The first term in \eqref{eq:lvae} measures the reconstruction error 
and the second term quantifies how well $q_\phi(z|x)$ matches the prior $p(z)$. The structure of the latent space heavily depends on this prior. As the KL divergence is non-negative, $-\Lvae$ lower-bounds the marginal $\log$-likelihood $\exe[\log p_\theta(x)]$ and is accordingly called the \emph{evidence lower bound (ELBO)}. 

There are several design choices available: (1) The prior distribution on the latent space, $p(z)$, (2) the family of approximate posterior distributions, $q_\phi(z|x)$, and (3) the decoder distribution, $p_\phi(x | z)$. Ideally, the approximate posterior should be flexible enough to match the intracable true posterior $p_\theta(z | x)$.  
As we will see later, there are many available options for these design choices, leading to various trade-offs in terms of the learned representation.

In practice, the first term in \eqref{eq:lvae} can be estimated from samples $z^{(i)} \sim q_\phi(z|x^{(i)})$ and gradients are backpropagated through the sampling operation using the \emph{reparametrization trick} \cite[Section~2.3]{kingma2013auto}, enabling minimization of \eqref{eq:lvae} via minibatch-stochastic gradient descent (SGD). Depending on the choice of $q_\phi(z|x)$ the second term can either be computed in closed form or estimated from samples. For the usual choice of $q_\phi(z|x) = \mc N(\mu_\phi(x), \diag(\sigma_\phi(x)))$, where $\mu_\phi(x)$ and $\sigma_\phi(x)$ are deterministic functions parametrized as neural networks, and $p(z) = \mc N(0,I)$ for which the KL-term in \eqref{eq:lvae} can be computed in closed form (more complicated choices of $p(z)$ rarely allow closed form computation). To this end, we will briefly discuss two ways in which one can measure distances between distributions. We will focus on intuition behind these techniques and provide pointers to detailed expositions.

\paragraph{Adversarial density-ratio estimation.}

\begin{figure}[t]
  \centering
    \begin{subfigure}[b]{.45\linewidth}
      \includegraphics[width=\textwidth]{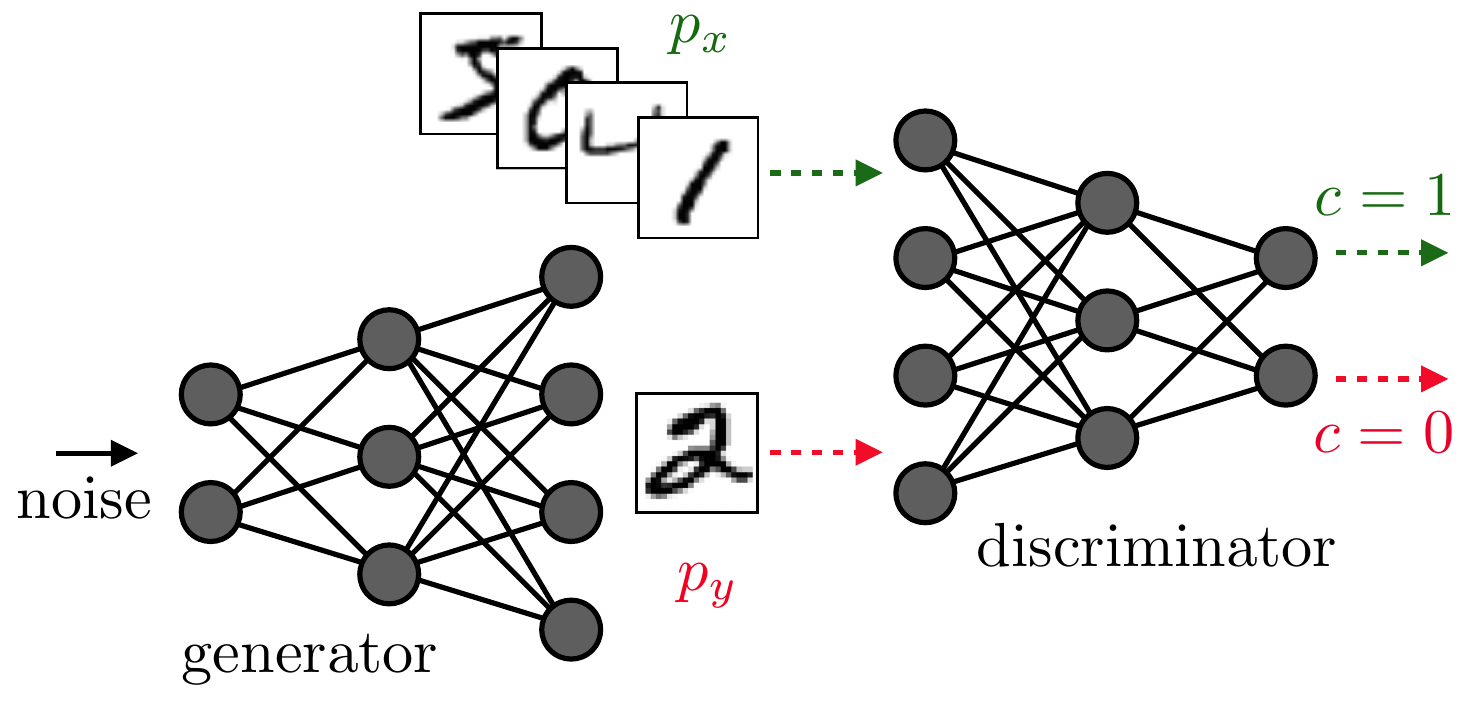}
      \subcaption{The main idea behind GANs.}
  \end{subfigure}\hfill
  \begin{subfigure}[b]{.5\linewidth}
    \includegraphics[width=\textwidth]{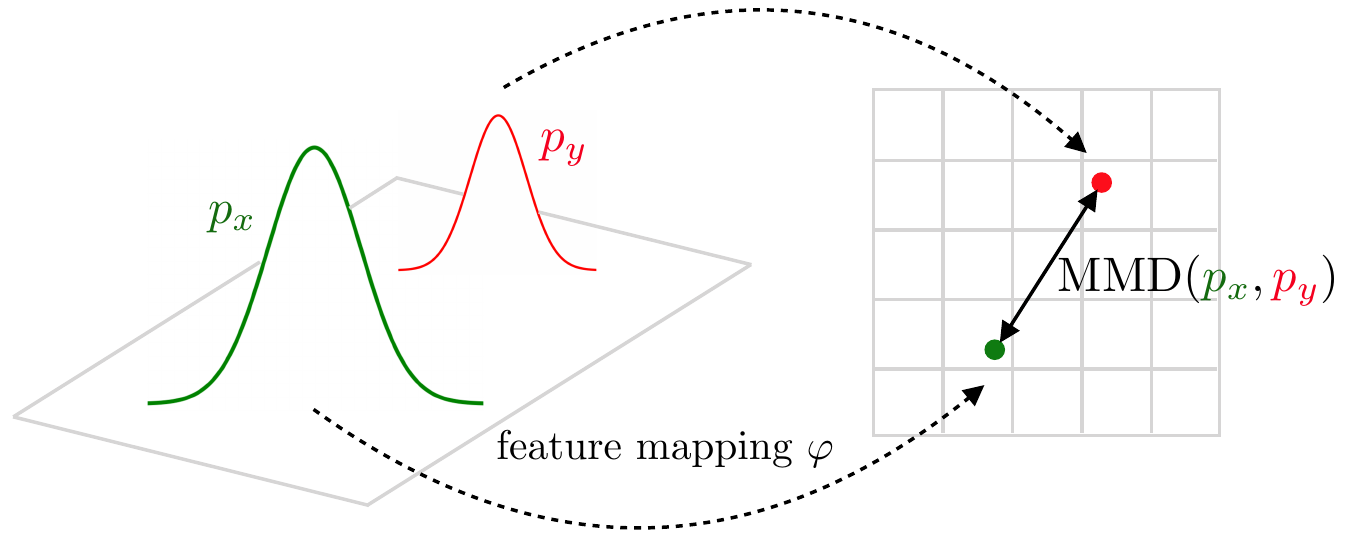}
    \subcaption{The main idea behind MMD.\label{fig:mmd}}
  \end{subfigure}
\caption{Adversarial density ratio estimation vs MMD.\label{fig:gansmmds} Figure (a): GANs use adversarial density ratio estimation to train a generative model, which can be seen as a two-player game: The discriminator tries to predict whether samples are real or generated, while the generator tries to deceive the discriminator by mimicking the distribution of the real samples. Figure (b): The MMD corresponds to the distance between mean feature embeddings.}
\end{figure}
Given a convex function $f$ for which $f(1)=0$, the $f$-divergence between $p_x$ and $p_y$ is defined as
\[
  D_f (p_x \| p_y)= \int f\left(\frac{p_x(x)}{p_y(x)}\right) p_y(x)dx.
\]
For example, the choice $f(t) = t \log t$ corresponds to $D_f(p_x\|p_y) = \dkl(p_x\|p_y)$. Given samples from $p_x$ and $p_y$ we can estimate the $f$-divergence using the density-ratio trick \cite{nguyen2010estimating,sugiyama2012density}, popularized recently through the generative adversarial network (GAN) framework \cite{goodfellow2014generative}. The trick is to express $p_x$ and $p_y$ as conditional distributions, conditioned on a label $c \in \{0, 1\}$, and reduce the task to binary classification. In particular, let $p_x(x) = p(x|c=1)$, $p_y(x) = p(x|c=0)$, and consider a discriminator $S_\eta$ trained to predict the probability that its input is a sample from distributions $p_x$ rather than $p_y$, i.e, predict $p(c=1|x)$. The density ratio can be expressed as
\begin{equation}
    \frac{p_x(x)}{p_y(x)} = \frac{p(x|c=1)}{p(x|c=0)} = \frac{p(c=1|x)}{p(c=0|x)} \approx \frac{S_\eta(x)}{1-S_\eta(x)},
\end{equation}
where the second equality follows from Bayes' rule under the assumption that the marginal class probabilities are equal. As such, given $N$ i.i.d. samples $\{x^{(i)}\}_{i=1}^N$ from $p_x$ and a trained classifier $S_\eta$ one can estimate the KL-divergence by simply computing
\[
  \dkl(p_x\|p_y) \approx \frac{1}{N} \sum_{i=1}^N \log\left(\frac{S_\eta(x^{(i)})}{1-S_\eta(x^{(i)})}\right).
\]
As a practical alternative, some approaches replace the KL term in \eqref{eq:lvae} with an arbitrary divergence (e.g., maximum mean discrepancy). Note, however, that the resulting objective does not necessarily lower-bound the marginal log-likelihood of the data.

\paragraph{Maximum mean discrepancy (MMD) \cite{gretton2012kernel}.} Intuitively, the distances between distributions are computed as distances between mean embeddings of features as illustrated in Figure~\ref{fig:mmd}. More formally, let $k\colon \mc X \to \mc X$ be a continuous, bounded, positive semi-definite kernel and $\mc H$ be the corresponding reproducing kernel Hilbert space, induced by the feature mapping $\varphi \colon \mc X \to \mc H$. Then, the MMD of distributions $p_x(x)$ and $p_y(y)$ is
\begin{equation}
    \mmd(p_x,p_y) = \|\mathbb{E}_{x\sim p_x}[\varphi(x)] - \mathbb{E}_{y \sim p_y}[\varphi(y)]\|_{\mc H}^2. \label{eq:mmdexact}
  \end{equation}
  For example, setting $\mc X = \mc H = \mathbb{R}^d$ and $\varphi(x) = x$, MMD reduces to the difference between the means, i.e., $\mmd(p_x, p_y) = \|\mu_{p_x} - \mu_{p_y}\|^2_2$. By choosing an appropriate mapping $\varphi$ one can estimate the divergence in terms of higher order moments of the distribution.

\paragraph{MMD vs $f$-divergences in practice.}
The MMD is known to work particularly well with multivariate standard normal distributions. It requires a sample size roughly on the order of the data dimensionality. When used as a regularizer (see Section~\ref{sec:regbased}), it generally allows for stable optimization. A disadvantage is that it requires selection of the kernel $k$ and its bandwidth parameter. In contrast, $f$-divergence estimators based on the density-ratio trick can in principle handle more complex distributions than MMD. However, in practice they require adversarial training which currently suffers from optimization issues. For more details consult \cite[Section~3]{tolstikhin2017wasserstein}.

\paragraph{Deterministic autoencoders.} Some of the methods we review rely on deterministic encoders and decoders. We denote by $D_\theta$ and $E_\phi$ the deterministic encoder and decoder, respectively. A popular objective for training an autoencoder is to minimize the $L_2$-loss, namely
\begin{equation} \label{eq:lae}
\Lae(\theta, \phi) = \frac{1}{2} \exe[\| x - D_\theta(E_\phi(x)) \|^2_2].
\end{equation}
If $E_\phi$ and $D_\theta$ are linear maps and the representation $z$ is lower-dimensional than $x$, \eqref{eq:lae} corresponds to principal component analysis (PCA), which leads to $z$ with decorrelated entries. Furthermore, we obtain \eqref{eq:lae} by removing the $\dkl$-term from $\Lvae$ in \eqref{eq:lvae} and using a deterministic encoding distribution $q_\phi(z|x)$ and a Gaussian decoding distribution $p_\theta(x|z)$. Therefore, the major difference between $\Lae$ and $\Lvae$ is that $\Lae$ does not enforce a prior distribution on the latent space (e.g., through a $\dkl$-term), and minimizing $\Lae$ hence does not yield a generative model.

\section{Regularization-based methods} \label{sec:regbased}

A classic approach to enforce some meta-prior on the latent representations $z \sim q_\phi(z|x)$ is to augment $\Lvae$ with regularizers that act on the approximate posterior $q_\phi(z|x)$ and/or the aggregate (approximate) posterior $q_\phi (z) = \exe[q_\phi(z|x)] = \frac{1}{N} \sum_{i=1}^N q_\phi(z|x^{(i)})$. 
The vast majority of recent work can be subsumed into an objective of the form
\begin{equation} \label{eq:regvaegeneral}
\Lvae(\theta,\phi) + \lambda_1 \exe[R_1(q_\phi(z|x))] + \lambda_2 R_2(q_\phi(z)),
\end{equation}
where $R_1$ and $R_2$ are regularizers and $\lambda_1,\lambda_2 > 0$ the corresponding weights. Firstly, we note a key difference between regularizers $R_1$ and $R_2$ is that the latter depends on the entire data set through $q_\phi (z)$. In principle, this prevents the use of mini-batch SGD to solve \eqref{eq:regvaegeneral}. In practice, however, one can often obtain good mini-batch-based estimates of $R_2(q_\phi(z))$. Secondly, the regularizers bias $\Lvae$ towards a looser (larger) upper bound on the negative marginal $\log$-likelihood. From this perspective it is not surprising that many approaches yield a lower reconstruction quality (which typically corresponds to a larger negative $\log$-likelihood). For deterministic autoencoders, there is no such concept as an aggregated posterior, so we consider objectives of the form $\Lae(\theta,\phi) + \lambda_1 \exe[R_1(E(x))]$.

In this section, we first review regularizers which can be computed in a fully unsupervised fashion (some of them optionally allow to include partial label information). Then, we turn our attention to regularizers which require supervision.

\begin{table}
  \centering
  \small
  \caption{\label{tab:infoth} Overview over different choices of the regularizers $R_1(q_\phi(z|x))$ and $R_2(q_\phi(z))$. The learning objective is specified in \eqref{eq:regvaegeneral}. Most approaches use a multivariate standard normal distribution as prior  (see Table~\ref{tab:infothcomp} in the appendix for more details). The last column (Y) indicates whether supervision is used: (\checkmark) indicates that labels are required,  while (O) indicates labels can optionally be used for (semi-) supervised learning. Note that some of the regularizers are simplified.\label{tab:disentanglement}}
  \vspace{2mm}
  \input{tables/disentanglement_table}
\end{table}

\subsection{Unsupervised methods targeting disentanglement and independence}
Disentanglement is a critical meta-prior considered by~\citet{bengio2013representation}. Namely, assuming the data is generated from a few statistically independent factors, uncovering those factors should be extremely useful for a plethora of downstream tasks. An example for (approximately) independent factors underlying the data are class, stroke  thickness, and rotation of handwritten digits in the MNIST data set. Other popular data sets are the CelebA face data set \cite{liu2015deep} (factors involve, e.g., hair color and facial attributes such as glasses), and synthetic data sets of geometric 2D shapes or rendered 3D shapes (e.g., 2D Shapes \cite{higgins2016beta}, 3D Shapes \cite{kim2018disentangling}, 3D Faces \cite{paysan20093d}, 3D Chairs \cite{aubry2014seeing}) for which the data generative process and hence the ground truth factors are known (see Figure~\ref{fig:disillustration} for an example). 

The main idea behind several recent works on disentanglement is to augment the $\Lvae$ loss with regularizers which encourage disentanglement of the latent variables $z$. Formally, assume that the data $x \sim p(x|v,w)$ depends on conditionally independent factors $v$, i.e., $p(v|x) = \prod_j p(v_j|x)$, and possibly conditionally dependent factors $w$. The goal is to augment $\Lvae$ such that the inference model $q_\phi(z|x)$ learns to predict $v$ and hence (partially) invert the data-generative process. 
\begin{figure}[t!]
  \input{figures/disentanglement_picture.tex}
  \caption{Schematic overview over different regularizers. Most approaches focus on regularizing the aggregate posterior and differ in the way the disagreement with respect to a prior is measured. More details are provided in Table~\ref{tab:infoth} and an in-depth discussion in Section~\ref{sec:regbased}.
  \label{fig:overview}}
\end{figure}
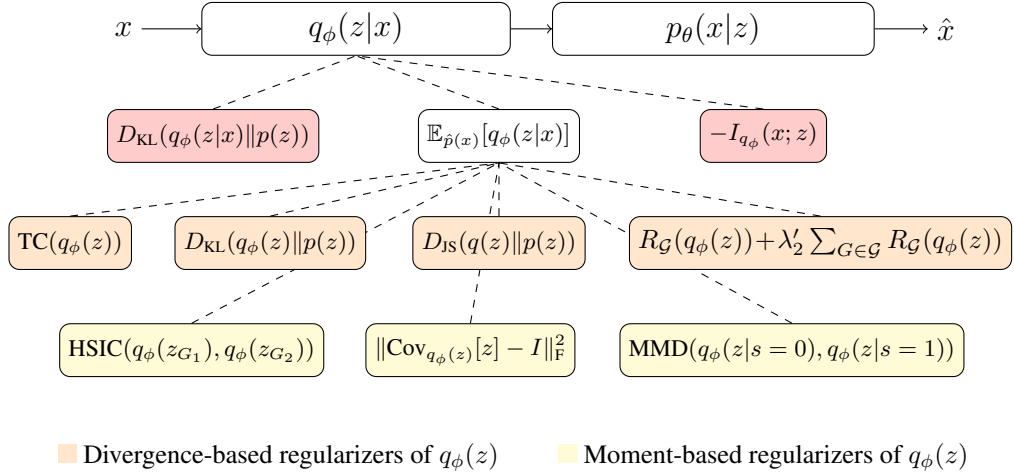

\paragraph{Metrics.}
Disentanglement quality of inference models is typically evaluated based on ground truth factors of variation (if available). Specifically, disentanglement metrics measure how predictive the individual latent factors are for the ground-truth factors, see, e.g., \cite{higgins2016beta, kim2018disentangling, kumar2018variational, eastwood2018framework, chen2018isolating, ridgeway2018learning}. While many authors claim
that their method leads to disentangled representations, it is unclear what the proper notion of disentanglement is and how effective these methods are in the unsupervised setting (see \cite{locatello2018challenging} for a large-scale evaluation). We therefore focus on the concept motivating each method rather than claims on how well each method disentangles the factors underlying the data.

\subsubsection{Reweighting the ELBO: $\beta$-VAE}
\citet{higgins2016beta} propose to weight the second term in \eqref{eq:lvae} (henceforth referred to as the $\dkl$-term) by a coefficient $\beta >1$,\footnote{\citet{higgins2016beta} also explore $0<\beta<1$ but discovers that this choice does not lead to disentanglement.} which can be seen as adding a regularizer equal to the $\dkl$-term with coefficient $\lambda_1 = \beta-1 >0$ to $\Lvae$
\begin{equation} \label{eq:betavae}
\mc L_{\beta\text{-VAE}}(\theta, \phi) = \Lvae(\theta, \phi) + \lambda_1 \exe[\dkl(q_\phi(z|x) \| p(z))].
\end{equation} This type of regularization encourages $q_\phi(z | x)$ to better match the factorized prior $p(z)$, which in turn constrains the implicit capacity of the latent representation $z\sim q_\phi(z|x)$ and encourages it be factorized.  \citet{burgess2018understanding} provide a through theoretical analysis of $\beta$-VAE based on the information bottleneck principle \cite{tishby2000information}. Further, they propose to gradually decrease the regularization strength until good quality reconstructions are obtained as a robust procedure to adjust the tradeoff between reconstruction quality and disentanglement (for a hard-constrained variant fo $\beta$-VAE).

\subsubsection{Mutual information of $x$ and $z$: FactorVAE, $\beta$-TCVAE, InfoVAE}
\citet{kim2018disentangling,chen2018isolating,zhao2017infovae} all propose regularizers motivated by the following decomposition of the second term in \eqref{eq:lvae}
\begin{equation} \label{eq:dkldecomp}
\exe[\dkl(q_\phi(z|x)\|p(z)] = I_{q_\phi}(x;z) + \dkl(q_\phi(z) \| p(z)),
\end{equation}
where $I_{q_\phi}(x; z)$ is the mutual information of $x$ and $z$ w.r.t. the distribution $q_\phi(x, z) = q_\phi(z|x) \hat p(x) =  \frac{1}{N} \sum_{i=1}^N q_\phi(z|x^{(i)}) \delta_{x^{(i)}}(x)$. The decomposition \eqref{eq:dkldecomp} was first derived by \citet{hoffman2016elbo}; an alternative derivation can be found in \citet[Appendix C]{kim2018disentangling}. 

\hlm{FactorVAE.} \citet{kim2018disentangling} observe that the regularizer in $\mc L_{\beta\text{-VAE}}$ encourages $q_\phi(z)$ to be factorized (assuming $p(z)$ is a factorized distribution) by penalizing the second term in \eqref{eq:dkldecomp}, but discourages the latent code to be informative by simultaneously penalizing the first term in \eqref{eq:dkldecomp}. To reinforce only the former effect, they propose to regularize $\Lvae$ with the total correlation $\tc(q_\phi(z))=\dkl(q_\phi(z) \| \textstyle\prod_j q_\phi(z_j))$ of~$q_\phi(z)$---a popular measure of dependence for multiple random variables \cite{watanabe1960information}. The resulting objective has the form
\begin{equation} \label{eq:factorvae}
\mc L_\text{FactorVAE}(\theta, \phi) = \Lvae(\theta, \phi) + \lambda_2 \tc(q_\phi(z))
\end{equation}
where the last term is the total correlation. To estimate it from samples, \citet{kim2018disentangling} rely on the density ratio trick \cite{nguyen2010estimating,sugiyama2012density} which involves training a discriminator (see Section~\ref{sec:prelim}). 

\hlm{$\beta$-TCVAE.} \citet{chen2018isolating} split up the second term in \eqref{eq:dkldecomp} as $\dkl(p(z) \| q_\phi(z)) = \dkl(q_\phi(z) \| \prod_j q_\phi(z_j)) + \sum_{j=1}^m \dkl(q_\phi(z_j)\|p(z_j))$ and penalize each term individually
\begin{equation} \label{eq:tcvae}
\mc L_{\beta\text{-TCVAE}}(\theta, \phi) = \Lvae(\theta, \phi) + \lambda_1 I_{q_\phi}(x;z) + \lambda_2 \tc(q_\phi(z)) + \lambda_2' \sum_j \dkl(q_\phi(z_j)\|p(z_j)). \nonumber
\end{equation}
However, they set $\lambda_1 = \lambda_2' = 0$ by default, effectively arriving at the same objective as FactorVAE in \eqref{eq:factorvae}. In contrast to FactorVAE, the TC-term is estimated using importance sampling.

\hlm{InfoVAE.} \citet{zhao2017infovae} start from an alternative way of writing $\Lvae$
\begin{equation} \label{eq:lvaeinfovae}
\Lvae(\theta, \phi) = \dkl(q_\phi(z) \| p(z)) + \exe[ \dkl(q_\phi(x|z) \| p_\theta (x|z))],
\end{equation}
where $q_\phi(x|z) = q_\phi(x, z)/p(z)$. Similarly to \cite{kim2018disentangling}, to encourage disentanglement, they propose to reweight the first term in \eqref{eq:lvaeinfovae} and to encourage a large mutual information between $z \sim q(z|x)$ and $x$ by adding a regularizer proportional to $I_{q_\phi}(x; z)$ to \eqref{eq:lvaeinfovae}. Further, by rearranging terms in the resulting objective, they arrive at
\begin{equation}\label{eq:infovae}
\mc L_\text{InfoVAE}(\theta, \phi) = \Lvae(\theta, \phi) + \lambda_1 \exe[ \dkl(q_\phi(z|x)\| p(z))] + \lambda_2 \dkl(q_\phi(z) \| p(z)).
\end{equation}
For tractability reasons, \citet{zhao2017infovae} propose to replace the last term in \eqref{eq:infovae} by other divergences such as Jensen-Shannon divergence (implemented as a GAN \cite{goodfellow2014generative}), Stein variational gradient \cite{liu2016stein}, or MMD \cite{gretton2012kernel} (see Section~\ref{sec:prelim}).
\begin{figure}[t]
  \centering
  \includegraphics[width=0.8\textwidth]{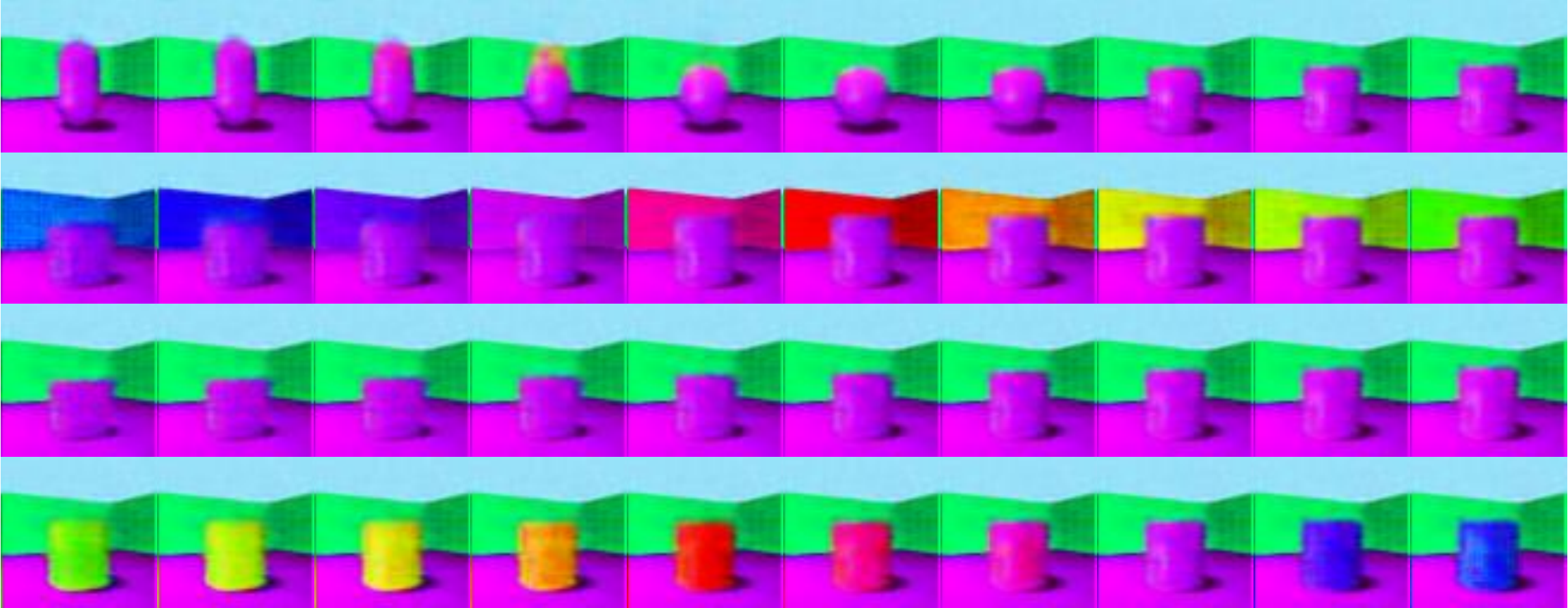}
  \caption{Latent space traversals obtained by varying one latent variable while holding all the others fixed (from left to right), for $\beta$-VAE \cite{higgins2016beta} (with $\beta=16$) trained on the 3D Shapes data set \cite{kim2018disentangling}. The variables shown correspond to the object shape, wall color, object height, and object color. Note that the other latent variables simultaneously vary multiple latent factors or are inactive for this model.}
  \label{fig:disillustration}
\end{figure}

\hlm{DIP-VAE.} \citet{kumar2018variational} suggest matching the moments of the aggregated posterior $q_\phi(z)$ to a multivariate standard normal prior $p(z)$ during optimization of $\Lvae$ to encourage disentanglement of the latent variables $z \sim q_\phi(z)$. Specifically, they propose to match the covariance of $q_\phi(z)$ and $\mc N(0,I)$ by penalizing their $\ell_2$-distance (which amounts to decorrelating the entries of $z \sim q_\phi(z)$) leading to the Disentangled Inferred Prior objective:
\begin{equation} \label{eq:dipvae}
\mc L_\text{DIP-VAE} (\theta, \phi) = \Lvae(\theta, \phi)
+ \lambda_2 \sum_{k \neq \ell} (\cov_{q_\phi(z)}[z])_{k,\ell}^2
+ \lambda_2' \sum_{k} ((\cov_{q_\phi(z)}[z])_{k,k}-1)^2.
\end{equation}
Noting that for the standard parametrization $q_\phi(z|x) = \mc N(\mu_\phi(x), \diag(\sigma_\phi(x)))$, $\cov_{q_\phi(z)}[z] = \sum_{i=1}^N \diag(\sigma_\phi(x_i)) + \cov_{\hat p(x)}[\mu_\phi(x)]$, $\sigma_\phi(x)$ only contributes to the diagonal of $\cov_{q_\phi(z)}[z]$, \citet{kumar2018variational} also consider a variant of $\mc L_\text{DIP-VAE}$ where $\cov_{q_\phi(z)}[z]$ in \eqref{eq:dipvae} is replaced by $\cov_{\hat p(x)}[\mu_\phi(x)]$.

\subsubsection{Independence between groups of latents: HSIC-VAE, HFVAE}

Groups/clusters, potentially involving hierarchies, is a structure prevalent in many data sets. It is therefore natural to take this structure into account when learning disentangled representations, as seen next.

\hlm{HSIC-VAE.} \citet{lopez2018information} leverage the Hilbert-Schmidt independence criterion (HSIC) \cite{gretton2005measuring} (cf. Section~\ref{sec:estimators}) to encourage independence between groups of latent variables, as
\begin{equation} \label{eq:hsicvae}
\mc L_\text{HSIC-VAE}(\theta, \phi) = \Lvae(\theta, \phi) + \lambda_2 \hsic(q_\phi(z_{G_1}), q_\phi(z_{G_2})),
\end{equation}
where $z_G=\{z_k\}_{k \in G}$ (an estimator of $\hsic$ is defined in \eqref{eq:hsicest} in Appendix~\ref{sec:estimators}). This is in contrast to the methods \cite{kim2018disentangling, chen2018isolating,zhao2017infovae, kumar2018variational} penalizing statistical dependence of all individual latent variables. In addition to controling (in)dependence relations of the latent variables, the HSIC can be used to remove sensitive information, provided as labels $s$ with the training data, from latent representation by using the regularizer $\hsic(q_\phi(z), p(s))$ (where $p(s)$ is estimated from samples) as extensively explored by~\citet{louizos2015variational} (see Section~\ref{sec:supmeth}).

\hlm{HFVAE.} Starting from the decomposition \eqref{eq:dkldecomp}, \citet{esmaeili2018structured} hierarchically decompose the $\dkl$-term in \eqref{eq:dkldecomp} into a regularization term of the dependencies between groups of latent variables $\mc G = \{G_k\}_{k=1}^{n_G}$ and regularization of the dependencies between the random variables in each group $G_k$. Reweighting different regularization terms allows to encourage different degrees of intra and inter-group disentanglement, leading to the following objective:
\begin{align} \
\mc L_\text{HFVAE} (\theta, \phi) &= \Lvae - \lambda_1 I_{q_\phi}(x;z) \nonumber\\
&\quad + \lambda_2 \left(- \mathbb \ex_{q_\phi (z)}\left[ \log \frac{p(z)}{\prod_{G\in \mc G} p(z_G)} \right] + \dkl(q_\phi(z)\|\textstyle\prod\limits_{G\in \mc G} q_\phi(z_G))\right) \nonumber \\
& \quad + \lambda_2' \sum_{G \in \mc G}
\left( -\mathbb \ex_{q_\phi (z_{G})}\left[ \log \frac{p(z_{G})}{\prod_{k \in G} p(z_k)} \right] + \dkl(q_\phi(z_{G})\|\textstyle\prod\limits_{k\in G} q_\phi(z_k)) \right). \label{eq:hfvae}
\end{align}
Here, $\lambda_1$ controls the mutual information between the data and latent variables, and $\lambda_2$ and $\lambda_2'$ determine the regularization of dependencies between groups and within groups, respectively, by penalizing the corresponding total correlation. Note that the grouping can be nested to introduce deeper hierarchies.

\subsection{Preventing the latent code from being ignored: PixelGAN-AE and VIB}

\hlm{PixelGAN-AE.} \citet{makhzani2017pixelgan} argue that, if $p_\theta(x|z)$ is not too powerful (in the sense that it cannot model the data distribution unconditionally, i.e., without using the latent code $z$) the term $I_{q_\phi}(x;z)$ in \eqref{eq:dkldecomp} and the reconstruction term in \eqref{eq:lvae} have competing effects: A small mutual information $I_{q_\phi}(x;z)$ makes reconstruction of $x^{(i)}$ from $q_\phi(z|x^{(i)})$ challenging for $p_\theta(x|z)$, leading to a large reconstruction error. Conversely, a small reconstruction error requires the code $z$ to be informative and hence $I_{q_\phi}(x;z)$ to be large. In contrast, if the decoder is powerful, e.g., a conditional PixelCNN \cite{van2016conditional}, such that it can obtain a small reconstruction error without relying on the latent code, the mutual information and reconstruction terms can be minimized largely independent, which prevents the latent code from being informative and hence providing a useful representation (this issue is known as the information preference property \cite{chen2016variational} and is discussed in more detail in Section~\ref{sec:structarch}). In this case, to encourage the code to be informative \citet{makhzani2017pixelgan} propose to drop the $I_{q_\phi}(x;z)$ term in \eqref{eq:dkldecomp}, which can again be seen as a regularizer
\begin{equation} \label{eq:pixelgan}
\mc L_\text{PixelGAN-AE}(\theta, \phi) = \Lvae(\theta, \phi) - I_{q_\phi}(x;z).
\end{equation}
The $\dkl$ term remaining in \eqref{eq:dkldecomp} after removing $I_{q_\phi}$ is approximated using a GAN. \citet{makhzani2017pixelgan} show that relying on $\mc L_\text{PixelGAN-AE}$ a powerful PixelCNN decoder can be trained while keeping the latent code informative. Depending on the choice of the prior (categorical or Gaussian), the latent code picks up information of different levels of abstraction, for example the digit class and writing style in the case of MNIST.

\hlm{VIB, information dropout.} \citet{alemi2016deep} and \citet{achille2018information} both derive a variational approximation of the information bottleneck objective \cite{achille2018information}, which targets learning a compact representation $z$ of some random variable $x$ that is maximally informative about some random variable $y$. In the special case, when $y=x$, the approximation derived in \cite{alemi2016deep} one obtains an objective equivalent to $\mc L_{\beta\text{-VAE}}$ in \eqref{eq:lvae} (c.f. \cite[Appendix~B]{alemi2016deep} for a discussion), whereas doing so for \cite{achille2018information} leads to
\begin{equation} \label{eq:infodrop}
\mc L_\text{InfoDrop} (\theta, \phi) = \Lvae(\theta, \phi) + \lambda_1 \exe[\dkl(q_\phi(z|x) \| p(z))] + \lambda_2 \tc(q_\phi(z)).
\end{equation}
\citet{achille2018information} derive (more) tractable expressions for \eqref{eq:infodrop} and establishe a connection to dropout for particular choices of $p(z)$ and $q_\phi(z|x)$. \citet{alemi2018fixing} propose an information-theoretic framework studying the representation learning properties of VAE-like models through a rate-distortion tradeoff. This framework recovers $\beta$-VAE but allows for a more precise navigation of the feasible rate-distortion region than the latter. \citet{alemi2018therml} further generalize the framework of \cite{alemi2016deep}, as discussed in Section~\ref{sec:critique}.

\subsection{Deterministic encoders and decoders: AAE and WAE}
Adversarial Autoencoders (AAEs) \cite{makhzani2015adversarial} turn a standard autoencoder into a generative model by imposing a prior distribution $p(z)$ on the latent variables by penalizing some statistical divergence $D_f$ between $p(z)$ and $q_\phi(z)$ using a GAN. Specifically, using the negative $\log$-likelihood as reconstruction loss, the AAE objective can be written as
\begin{equation}\label{eq:aae}
\mc L_\text{AAE}(\theta, \phi)
= \exe[ \ex_{q_\phi(z|x)}[- \log p_\theta(x|z)]] + \lambda_2 D_f(q(z)\|p(z)).
\end{equation}
In all experiments in \cite{makhzani2015adversarial} encoder and decoder are taken to be deterministic, i.e., $p(x|z)$ and $q(z|x)$ are replaced by $D_\theta$ and $E_\phi$, respectively, and the negative $\log$-likelihood in \eqref{eq:aae} is replaced with the standard autoencoder loss $\Lae$. The advantage of implementing the regularizer $\lambda_2 D_f$ using a GAN is that any $p(z)$ we can sample from, can be matched. This is helpful to learn representations: For example for MNIST, enforcing a prior that involves both categorical and Gaussian latent variables is shown to disentangle discrete and continuous style information in unsupervised fashion, in the sense that the categorical latent variables model the digit index and continuous random variables the writing style. Disentanglement can be improved by leveraging (partial) label information, regularizing the cross-entropy between the categorical latent variables and the label one-hot encodings. Partial label information also allows to learn a generative model for digits with a Gaussian mixture model prior, with every mixture component corresponding to one digit index.

\subsection{Supervised methods: VFAEs, FaderNetworks, and DC-IGN} \label{sec:supmeth}

\paragraph{VFAE.} Variational Fair Autoencoders (VFAEs) \cite{louizos2015variational} assume a likelihood of the form $p_\theta(x | z, s)$, where $s$ models (categorical) latent factors one wants to remove (for example sensitive information), and $z$ models the remaining latent factors. By using an approximate posterior of the form $q_\phi(z | x, s)$ and by imposing factorized prior $p(z)p(s)$ one can encourage independence of $z \sim q_\phi(z | x, s)$ from $s$. However, $z$ might still contain information about $s$, in particular in the (semi-) supervised setting where $z$ encodes label information $y$ that might be correlated with $s$, and additional factors of variation $z'$, i.e., $z \sim p_\theta(z| z', y)$ (this setup was first considered in \cite{kingma2014semi}; see Section~\ref{sec:structarch}). To mitigate this issue, \citet{louizos2015variational} propose to add an MMD-based regularizer to $\Lvae$, encouraging independence between $q(z|s = k)$ and $q(z|s=k')$, i.e.,
\begin{equation} \label{eq:vfae}
\mc L_\text{VFAE}(\theta, \phi) = \Lvae + \lambda_2 \sum_{\ell=2}^{K} \mmd(q_\phi(z|s=\ell), q_\phi(z|s=1)),
\end{equation}
where $q_\phi(z|s = \ell) =  \sum_{i \colon s^{(i)} = \ell} \frac{1}{|\{i \colon s^{(i)} = \ell\}|} q_\phi(z| x^{(i)}, s^{(i)})$. 
To reduce the computational complexity of the MMD the authors propose to use random Fourier features \cite{rahimi2008random}. \citet{lopez2018information} also consider the problem of censoring side information, but use the $\hsic$ regularizer instead of MMD. In contrast to MMD, the $\hsic$ is amenable to side information $s$ of a non-categorical distribution. Furthermore, it is shown in \citet[Appendix E]{lopez2018information} that VFAE and HSIC are equivalent to censoring in case $s$ is a binary random variable.

\hlm{Fader Networks.} A supervised method similar to censoring outlined above was explored by \citet{lample2017fader} and \citet{hadad2018two}. Given data $\{x^{(i)}\}_{i=1}^N$ (e.g., images of faces) and corresponding binary attribute information $\{y^{(i)}\}_{i=1}^N$ (e.g., facial attributes such as hair color or whether glasses are present; encoded as binary vector in $\{0,1\}^K$), the encoder of a FaderNetwork \cite{lample2017fader} is adversarially trained to learn a feature representation $z = E_\phi(x)$ invariant to the attribute values, and the decoder $D_\theta(y,z)$ reconstructing the original image from $z$ and $y$. The resulting model is able to manipulate the attributes of a testing image (without known attribute information) by setting the entries of $y$ at the input of $D_\theta$ as desired. In particular, it allows for continuous control of the attributes (by choosing non-integer attribute values in $[0,1]$).

To make $z = E_\phi(x)$ invariant to $y$ a discriminator $P_\psi(y | z)$ predicting the probabilities of the attribute vector $y$ from $z$ is trained  concurrently with $E_\phi, D_\theta$ to maximize the $\log$-likelihood $\mc L_\text{dis} (\psi) = \exey[\log P_\psi(y | E_\phi(x))]$. This discriminator is used adversarially in the training of $E_\phi$, $D_\theta$ encouraging $E_\phi$ to produce a latent code $z$ from which it is difficult to predict $y$ using $P_\psi$ as
\begin{equation}\label{eq:fadernets}
\mc L_\text{Fader}(\theta, \phi) = \exey \left[ \frac{1}{2} \| x - D_\theta(y, E_\phi(x)) \|^2_2 - \lambda_1 \log P_\psi(1-y | E_\phi(x)) \right],
\end{equation}
i.e., the regularizer encourages $E_\phi$ to produce codes for which $P_\psi$ assigns a high likelihood to incorrect attribute values.

\citet{hadad2018two} propose a method similar to FaderNetworks that first separately trains an encoder $z'=E'_{\phi'}(x)$ jointly with a classifier to predict $y$. The code produced by $E'_{\phi'}$ is then concatenated with that produced by a second encoder $E''_{\phi''}$ and fed to the decoder $D_\theta$. $E''_{\phi''}$ and $D_\theta$ are now jointly trained for reconstruction (while keeping $\phi'$ fixed) and the output of $E''_{\phi''}$ is regularized as in \eqref{eq:fadernets} to ensure that $z''=E''_{\phi''}$ and $z'=E'_{\phi'}$ are disentangled. While the model from \cite{hadad2018two} does not allow fader-like control of attributes, it provides a representation that facilitates swapping and interpolation of attributes, and can be use for retrieval. Note that in contrast to all previously discussed methods, both of these techniques do not provide a mechanism for unconditional generation.

\paragraph{DC-IGN.} \citet{kulkarni2015deep} assume that the training data is generated by an interpretable, compact graphics code and aim to recover this code from the data using a VAE. Specifically, they consider data sets of rendered object images for which the underlying graphics code consists of extrinsic latent variables---object rotation and light source position---and intrinsic latent variables, modeling, e.g., object identity and shape. Assuming supervision in terms of which latent factors are active (relative to some reference value), a representation disentangling intrinsic and the different extrinsic latent variables is learned by optimizing $\Lvae$ on different types of mini-batches (which can be seen as implicit regularization): Mini-batches containing images for which all but one of the extrinsic factors are fixed, and mini-batches containing images with fixed extrinsic factors, but varying intrinsic factors. During the forward pass, the latent variables predicted by the encoder corresponding to fixed factors are replaced with the mini-batch average to force the decoder to explain all the variance in the mini-batch through the varying latent variables. In the backward step, gradients are passed through the latent space ignoring the averaging operation. This procedure allows to learn a disentangled representation for rendered 3D faces and chairs that allow to control extrinsic factors similarly as in a rendering engine. The models generalize to unseen object identities.

\section{Factorizing the encoding and decoding distributions} \label{sec:structarch}

\begin{figure}
  \centering
  \begin{subfigure}[b]{.47\linewidth}
    \centering
    \includegraphics[clip, trim=3.2cm 4.5cm 4cm 18cm, width=1\textwidth]{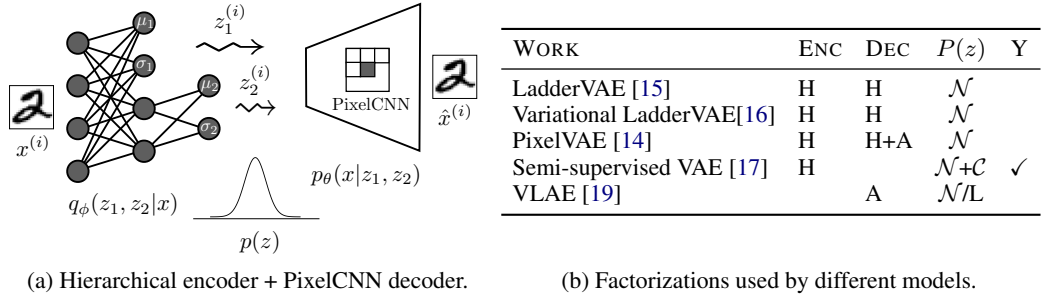}
    \subcaption{Hierarchical encoder + PixelCNN decoder.}
    \end{subfigure}
  \begin{subfigure}[b]{.51\linewidth}
    \centering%
    \small
    \input{tables/structured_arch.tex}
    \par\vspace{0.5cm}
    \subcaption{Factorizations used by different models.}
   \end{subfigure}
\caption{
 \label{tab:encdecdistr} Figure (a) shows an example VAE with hierarchical encoding distribution and PixelCNN decoding distribution. Figure (b) gives an overview of factorizations used by different models. We indicate the structure of the encoding (ENC) and decoding (DEC) distribution as follows: (H) hierarchical, (A) autoregressive, (default) fully connected or convolutional feed-feed forward neural network). We indicate the prior distribution as follows: ($\N$) multivariate standard Normal, ($\C$) categorical, (M) mixture distribution, (G) graphical model, (L) learned prior. The last column (Y) indicates whether supervision is used.
}
\label{fig:test}
\end{figure}

Besides regularization, another popular way to impose a meta-prior is factorizing the encoding and/or decoding distribution in a certain way (see Figure~\ref{tab:encdecdistr} for an overview). This translates directly or indirectly into a particular choice of the model class/network architecture underlying these distributions. Concrete examples are hierarchical architectures and architectures with constrained receptive field. This can be seen as hard constraints on the learning problem, rather than regularization as discussed in the previous section. While this is not often done in the literature, one could obviously combine a specific structured model architecture with some regularizer, for example to learn a disentangled hierarchical representation. Choosing a certain model class/architecture is not only interesting from a representation point of view, but also from a generative modeling perspective. Indeed, certain model classes/architectures allow to better optimize $\Lvae$ ultimately leading to a better generative model.

\paragraph{Semi-supervised VAE.} \citet{kingma2014semi} harness the VAE framework for semi-supervised learning. Specifically, in the ``M2 model'', the latent code is divided into two parts $z$ and $y$ where $y$ is (typically discrete) label information observed for a subset of the training data. More specifically, the inference model takes the form $q_\phi(z, y | x) = q_\phi(z |y, x)q_\phi(y | x)$, i.e., there is a hierarchy between $y$ and $z$. During training, for samples $x^{(i)}$ for which a label $y^{(i)}$ is a available, the inference model is conditioned on $y$ (i.e., $q_\phi(z|y, x)$) and $\Lvae$ is adapted accordingly, and for samples without label, the label is inferred from $q_\phi(z, y | x)$. This model hence effectively disentangles the latent code into two parts $y$ and $z$ and allows for semi-supervised classification and controlled generation by holding one of the factors fixed and generating the other one. This model can optionally be combined with an additional model learned in unsupervised fashion to obtain an additional level of hierarchy (termed ``M1 + M2 model'' in \cite{kingma2014semi}).

\paragraph{VLAE.} 
Analyzing the VAE framework through the lens of Bits-Back coding \cite{hinton1993keeping,honkela2004variational}, \citet{chen2016variational} identify the so-called information preference property: The second term in $\Lvae$ \eqref{eq:lvae} encourages the latent code $z \sim q_\phi(z|x)$ to only store the information that cannot be modeled locally (i.e., unconditionally without using the latent code) by the decoding distribution $p_\theta(x|z)$. As a consequence, when the decoding distribution is a powerful autoregressive model such as conditional PixelRNN \cite{van2016pixel} or PixelCNN \cite{van2016conditional} the latent code will not be used to encode any information and $q_\phi(z|x)$ will perfectly match the prior $p(z)$, as previously observed by many authors. While this not necessarily an issue in the context of generative modeling (where the goal is to maximize testing $\log$-likelihood), it is problematic from a representation learning point of view as one wants the latent code $z\sim q_\phi(z|x)$ to store meaningful information. To overcome this issue, \citet{chen2016variational} propose to adapt the structure of the decoding distribution $p_\theta(x|z)$ such that it cannot model the information one would like $z$ to store, and term the resulting model variational lossy autoencoder (VLAE). For example, to encourage $z$ to capture global high-level information, while letting $p_\theta(x|z)$ model local information such as texture, one can use an autoregressive decoding distribution with a limited local receptive field $p_\theta(x|z) = \prod_j p_\theta(x_j| z, x_{W(j)})$, where $W(j)$ is a window centered in pixel $j$, that cannot model long-range spatial dependencies. 
Besides the implications of the  information preference property for representation learning, \citet{chen2016variational} also explore the orthogonal direction of using a learned prior based on autoregressive flow \cite{kingma2016improved} to improve generative modeling capabilities of VLAE. 

\paragraph{PixelVAE.} PixelVAEs \cite{gulrajani2016pixelvae} use a VAE with feed-forward convolutional encoder and decoder, combining the decoder with a (shallow) conditional PixelCNN \cite{van2016conditional} to predict the output probabilities. Furthermore, they employ a hierarchical encoder and decoder structure with multiple levels of latent variables. In more detail, the encoding and decoding distributions are factorized as $q_\phi(z_1, \ldots, z_L|x) = q_\phi(z_1|x) \ldots q_\phi(z_L|x)$ and $p_\theta(x, z_1, \ldots, z_L) = p_\theta(x|z_1) p_\theta(z_1|z_2)\ldots p_\theta(z_{L-1}|z_L)p(z_L)$. Here, $z_1,\ldots,z_L$ are groups of latent variables (rather than individual entries of $z$), the $q_\phi(z_j|x)$ are parametric distributions (typically Gaussian with diagonal covariance matrix) whose parameters are predicted from different layers of the same CNN (with layer index increasing in $j$), $p_\theta(x|z_1)$ is a conditional PixelCNN, and the factors in $p_\theta(z_1|z_2)\ldots p_\theta(z_{L-1}|z_L)$ are realized by a feed-forward convolutional networks. From a representation learning perspective, this approach leads to the extraction of high- and low-level features on one hand, allowing for controlled generation of local and global structure, and on the other hand results in better clustering of the codes according to classes in the case of multi-class data. From a generative modeling perspective, this approach obtains testing likelihood competitive with or better than computationally more complex (purely autoregressive) PixelCNN and PixelRNN models. Only $L=2$ stochastic layers are explored experimentally.

\paragraph{LadderVAE.} In contrast to PixelVAEs, Ladder VAEs (LVAEs) \cite{sonderby2016ladder} perform top-down inference, i.e., the encoding distribution is factorized as $q_\phi(z|x) = q_\phi(z_L|x) \prod_{j=1}^{L-1} q_\phi(z_j|z_{j+1})$, while using the same factorization for $p_\theta(x|z)$ as PixelVAE (although employing a simple factorized Gaussian distribution for $p_\theta(x|z_1)$ instead of a PixelCNN). The $q_\phi(z_j |z_{j+1})$ are parametrized Gaussian distributions whose parameters are inferred top-down using a precision-weighted combination of (i) bottom-up predictions from different layers of the same feed-forward encoder CNN (similarly as in PixelVAE) with (ii) top-down predictions obtained by sampling from the hierarchical distribution $p_\theta(z) = p_\theta(z_1|z_2)\ldots p_\theta(z_{L-1}|z_L)p(z_L)$ (see \cite[Figure~1b]{sonderby2016ladder} for the corresponding graphical model representation). When trained with a suitable warm-up procedure, LVAEs are capable of effectively learning deep hierarchical latent representations, as opposed to hierarchical VAEs with bottom-up inference models which usually fail to learn meaningful representations with more than two levels (see \cite[Section 3.2]{sonderby2016ladder}).

\paragraph{Variational Ladder AutoEncoders.} Yet another approach is taken by Variational Ladder autoencoders (VLaAEs) \cite{zhao2017learning}: While no explicit hierarchical factorization of $p(z)$ in terms of the $z_j$ is assumed, $p_\theta(z_1|z_2,\ldots z_L)$ is implemented as a feed-forward neural network, implicitly defining a top-down hierarchy among the $z_j$ by taking the $z_j$ as inputs on different layers, with the layer index proportional to $j$. $p_\theta(x|z_1)$ is set to a fixed variance factored Gaussian whose mean vector is predicted from $z_1$. For the encoding distribution $q_\phi(z|x)$ the same factorization and a similar implementation as that of PixelVAE is used. Implicitly encoding a hierarchy into $p_\theta(z_1|z_2,\ldots z_L)$ rather than explicitly as by PixelVAE and LVAE avoids the difficulties described by \cite{sonderby2016ladder} involved with training hierarchical models with more than two levels of latent variables. Furthermore, \citet{zhao2017learning} demonstrate that this approach leads to a disentangled hierarchical representation, for instance separating stroke width, digit width and tilt, and digit class, when applied to MNIST.

Finally, \citet{bachman2016architecture} and \citet{kingma2016improved} explore hierarchical factorizations/architectures mainly to improve generative modeling performance (in terms of testing $\log$-likelihood), rather than exploring it from a representation learning perspective.
\section{Structured prior distribution} \label{sec:structprior}

\begin{figure}
  \centering
  \begin{subfigure}{0.54\linewidth}
    \centering
    \includegraphics[clip, trim=3.2cm 12.5cm 3.7cm 11cm, width=1\textwidth]{figures/rep_learning_illustrations}
\subcaption{VAE with a multimodal continuous or discrete prior.}
\end{subfigure}~
\begin{subfigure}{0.45\linewidth}
    \centering%
    \small
    \input{tables/structured_prior.tex}
    \par\vspace{0.3cm}
\subcaption{Priors employed by different models.}
\end{subfigure}
\caption{ \label{tab:priors} 
Figure (a) shows an example of a VAE with with a multimodal continuous or discrete prior (each prior gives rise to a different model). Figure (b) gives an overview of the priors employed by different models.  
We indicate the structure of the encoding (ENC) and decoding (DEC) distribution as follows: (H) hierarchical, (A) autoregressive, (default) fully connected or convolutional feed-feed forward neural network. We indicate the prior distribution as follows: ($\N$) multivariate standard Normal, ($\C$) categorical, (M) mixture distribution, (G) graphical model, (L) learned prior. The last column (Y) indicates whether supervision is used: (\checkmark) indicates that labels are required.
\label{tab:disentanglement}}
\end{figure}

Instead of choosing the encoding distribution, one can also encourage certain meta-priors by directly choosing the prior distribution $p(z)$ of the generative model. For example, relying on a prior involving discrete and continuous random variables encourages them to model different types of factors, such as the digits and the writing style, respectively, in the MNIST data set, which can be seen as a form of clustering. This is arguably the most explicit way to shape a representation, as the prior directly acts on its distribution.

\subsection{Graphical model prior}

\paragraph{SVAE.}
One of the first attempts to learn latent variable models with structured prior distributions using the VAE framework is \cite{johnson2016composing}. Concretely, the latent distribution $p(z)$ with general graphical model structure can capture discrete mixture models such as Gaussian mixture models, linear dynamical systems, and switching linear dynamical systems, among others. Unlike many other VAE-based works, \citet{johnson2016composing} rely on a fully Bayesian framework including hyperpriors for the likelihood/decoding distribution and the structured latent distribution. While such a structured $p(z)$ allows for efficient inference (e.g., using message passing algorithms) when the likelihood is an exponential family distribution, it becomes intractable when the decoding distribution is parametrized through a neural network as commonly done in the VAE framework, the reason for which the latter includes an approximate posterior/encoding distribution. To combine the tractability of conjugate graphical model inference with the flexibility of VAEs, \citet{johnson2016composing} employ inference models that output conjugate graphical model potentials \cite{wainwright2008graphical} instead of the parameters of the approximate posterior distribution. In particular, these potentials are chosen such that they have a form conjugate to the exponential family, hence allowing for efficient inference when combined with the structured $p(z)$. The resulting algorithm is termed structured VAE (SVAE). Experiments show that SVAE with a Gaussian mixture prior learns a generative model whose latent mixture components reflect clusters in the data, and SVAE with a switching linear dynamical system prior learns a representation that reflects behavior state transitions in motion recordings of mouses.

\citet{narayanaswamy2017learning} consider latent distributions with graphical model structure similar to \cite{johnson2016composing}, but they also incorporate partial supervision for some of the latent variables as \cite{kingma2014semi}. However, unlike \citet{kingma2014semi} which assumes a posterior of the form $q_\phi(z, y | x) = q_\phi(z |y, x)q_\phi(y | x)$, they do not assume a specific factorization of the partially observed latent variables $y$ and the unobserved ones $z$ (neither for $q_\phi(z,y|x)$ nor for the marginals $q_\phi(z|x)$ and $q_\phi(y|x)$), and no particular distributional form of $q_\phi(z|x)$ and $q_\phi(y|x)$. To perform inference for $q_\phi(z,y|x)$ with arbitrary dependence structure, \citet{narayanaswamy2017learning} derive a new Monte Carlo estimator. The proposed approach is able to disentangle digit index and writing style on MNIST with partial supervision of the digit index (similar to \cite{kingma2014semi}). Furthermore, this approach can disentangle identity and lighting direction of face images with partial supervision assuming the product of categorical and continuous distribution, respectively, for the prior (using the the Gumbel-Softmax estimator \cite{jang2016categorical, maddison2016concrete} to model the categorical part in the approximate posterior).

\subsection{Discrete latent variables}

\paragraph{JointVAE.} JointVAE \cite{dupont2018joint} equips the $\beta$-VAE framework with heterogeneous latent variable distributions by concatenating continuous latent variables $z$ with discrete ones $c$ for improved disentanglement of different types of latent factors. The corresponding approximate posterior is factorized as $q_\phi(c|x)q_\phi(z|x)$ and the Gumbel-Softmax estimator \cite{jang2016categorical, maddison2016concrete} is used to obtain a differentiable relaxation of the categorical distribution $q_\phi(c|x)$. The regularization strength $\lambda_1$ in the (a constrained variant of) $\beta$-VAE objective \eqref{eq:betavae} is gradually increased during training, possibly assigning different weights to the regularization term corresponding to the discrete and continuous random variables (the regularization term in \eqref{eq:betavae} decomposes as $\dkl(q_\phi(z|x)q_\phi(c|x)\| p(z)p(c)) = \dkl(q_\phi(z|x)\| p(z)) + \dkl(q_\phi(c|x)\|p(c))$). Numerical results (based on visual inspection) show that the discrete latent variables naturally model discrete factors of variation such as digit class in MNIST or garment type in Fashion-MNIST and hence disentangle such factors better than models with continuous latent variables only.

\paragraph{VQ-VAE.} \citet{van2017neural} realize a VAE with discrete latent space structure using vector quantization, termed VQ-VAE. Each latent variable $z_j$ is taken to be a categorical random variable with $K$ categories, and the approximate posterior $q_\phi(z_j|x)$ is assumed deterministic. Each category is associated with an embedding vector $e_k \in \mathbb R^D$. The embedding operation induces an additional latent space dimension of size $D$. For example, if the latent representation $z$ is an $M \times M \times 1$ feature map, the embedded latent representation $\tilde z$ is a $M\times M \times D$ feature map. The distribution $q_\phi(\tilde z_j|x)$ is implemented using a deterministic encoder network $E_\phi(x)$ with $D$-dimensional output, quantized w.r.t. the embedding vectors $\{e_k\}_{k=1}^K$. In summary, we have
\begin{equation} \label{eq:vqvae}
q_\phi(\tilde z_j = e_k|x) =
\begin{cases}
1 \quad \text{if } k = \arg \min_{\ell} \|E_\phi(x) - e_\ell\|, \\
0 \quad \text{otherwise.}
\end{cases}
\end{equation}
The embeddings $e_k$ can be learned individually for each latent variable $z_j$, or shared for the entire latent space. Assuming a uniform prior $p(z)$, the second term in $\Lvae$ \eqref{eq:lvae} evaluates to $\log K$ as a consequence of $q_\phi(z|x)$ being deterministic and can be discarded during optimization. To backpropagate gradients through the non-differentiable operation \eqref{eq:vqvae} a straight-through type estimator \cite{bengio2013estimating} is used. The embedding vectors $e_k$, which do not receive gradients as a consequence of using a straight-through estimator, are updated as the mean of the encoded points $E_\phi(x^{(i)})$ assigned to the corresponding category $k$ as in (mini-batch) $k$-means.

VQ-VAE is shown to be competitive with VAEs with continuous latent variables in terms of testing likelihood. Furthermore, when trained on speech data, VQ-VAE learns a rudimentary phoneme-level language model in a completely unsupervised fashion, which can be used for controlled speech generation and phoneme classification.

Many other works explore learning (variational) autoencoders with (vector-)quantized latent representation with a focus on generative modeling \cite{mnih2014neural, mnih2016variational, jang2016categorical, maddison2016concrete} and compression \cite{agustsson2017soft}, rather than representation learning.

\section{Other approaches} \label{sec:otherapp}

\paragraph{Early approaches.} Early approaches to learn abstract representations using autoencoders include stacking single-layer autoencoders \cite{bengio2007greedy} to build deep architectures and imposing a sparsity prior to the latent variables \cite{ranzato2007efficient}. Another way to achieve abstraction is to require the representation to be robust to noise. Such a representation can be learned using denoising autoencoders \cite{vincent2008extracting}, i.e., autoencoders trained to reconstruct clean data points from a noisy version. For a broader overview over early approaches we refer to \cite[Section~7]{bengio2013representation}.

\paragraph{Sequential data.} There is a considerable number of recent works leveraging (variational) autoencoders and the techniques similar to those outlined in Sections~\ref{sec:regbased}--\ref{sec:structprior} to learn representations of sequences. \citet{yingzhen2018disentangled} partition the latent code of a VAE into subsets of time varying and time invariant variables (resulting in a particular factorization of the approximate posterior) to learn a representation disentangling content and pose/identity in video/audio sequences. \citet{hsieh2018learning} use a similar partition of the latent code, but additionally allow the model to decompose the input into different parts, e.g., modelling different moving objects in a video sequence. Somewhat related, \citet{villegas2017decomposing, denton2017unsupervised, fraccaro2017disentangled} propose autoencoder models for video sequence prediction with separate encoders disentangling the latent code into pose and content. \citet{hsu2017unsupervised} develop a hierarchical VAE model to learn interpretable representations of speech recordings. \citet{fortuin2018deep} combine a variation of VQ-VAE with self-organizing maps to learn interpretable discrete representations of sequences. Further, VAEs for sequences are also of great interest in the context of natural language processing, in particular with autoregressive encoders/decoders and discrete latent representations, see, e.g., \cite{bowman2016generating, hu2017toward, serban2017hierarchical} and references therein.

\paragraph{Using a discriminator in pixel space.} An alternative to training a pair of probabilistic encoder $q_\phi(z|x)$ and decoder $p_\theta(x|z)$ to minimize a reconstruction loss is to learn $\phi,\theta$ by matching the joint distributions $p_\theta(x|z)p(z)$ and $q_\phi(z|x)\hat p(x)$. To achieve this, adversarially learned inference (ALI) \cite{dumoulin2016adversarially} and bidirectional GAN (BiGAN) \cite{donahue2016adversarial} leverage the GAN framework, learning $p_\theta(x|z)$, $q_\phi(z|x)$ jointly with a discriminator to distinguish between samples drawn from the two joint distributions. While this approach yields powerful generative models with latent representations useful for downstream tasks, the reconstructions are less faithful than for autoencoder-based models. \citet{li2017alice} point out a non-identifiability issue inherent with the distribution matching problem underlying ALI/BiGAN, and propose to penalize the entropy of the reconstruction conditionally on the code.

\citet{chen2016infogan} augment a standard GAN framework \cite{goodfellow2014generative} with a mutual information term between the generator output and a subset of latent variables, which proves effective in learning disentangled representations. Other works regularize the output of (variational) autoencoders with a GAN loss. Specifically, \citet{larsen2015autoencoding,rosca2017variational} combine VAE with standard GAN  \cite{goodfellow2014generative}, and \citet{tschannen2018deep} equip AAE/WAE with a Wasserstein GAN loss \cite{arjovsky17wasserstein}. While \citet{larsen2015autoencoding} investigate the representation learned by their model, the focus of these works is on improving the sample quality of VAE and AAE/WAE. \citet{mathieu2016disentangling} rely on a similar setup as \cite{larsen2015autoencoding}, but use labels to learn disentangled representations.

\paragraph{Cross-domain disentanglement.} Image-to-image translation methods \cite{isola2017image,zhu2017unpaired} (translating, e.g., semantic label maps into images) can be implemented by training  encoder-decoder architectures to translate between two domains (i.e., in both directions) while enforcing the translated data to match the respective domain distribution. While this task as such does not a priori encourage learning of meaningful representation, adding appropriate pressure does: Sharing parts of the latent representation between the translation networks \cite{liu2017detach,lee2018diverse,gonzalez2018image} and/or combining domain specific and shared translation networks \cite{liu2018unified} leads to disentangled representations.

\section{Rate-distortion tradeoff and usefulness of representation} \label{sec:critique}
\begin{figure}[t]
 \centering
 \begin{subfigure}[b]{.4\linewidth}
   \includegraphics[width=\textwidth]{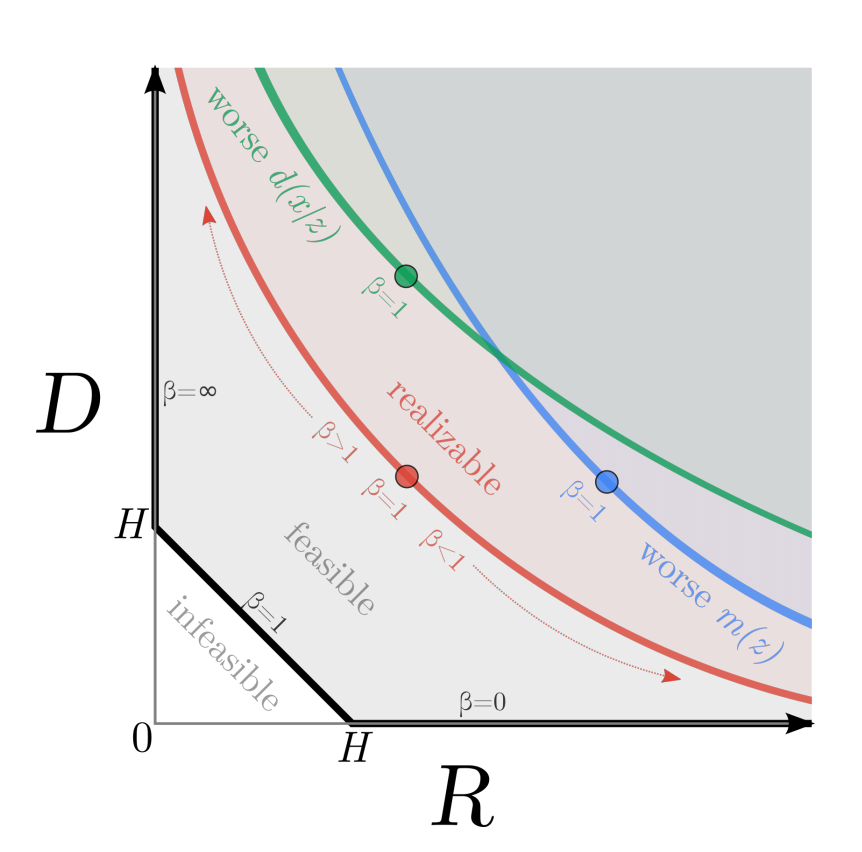}
   \caption{Rate-distortion ($R$-$D$) tradeoff of \cite{alemi2018fixing}.}
 \end{subfigure} \qquad
 \begin{subfigure}[b]{.4\linewidth}
   \includegraphics[width=\textwidth]{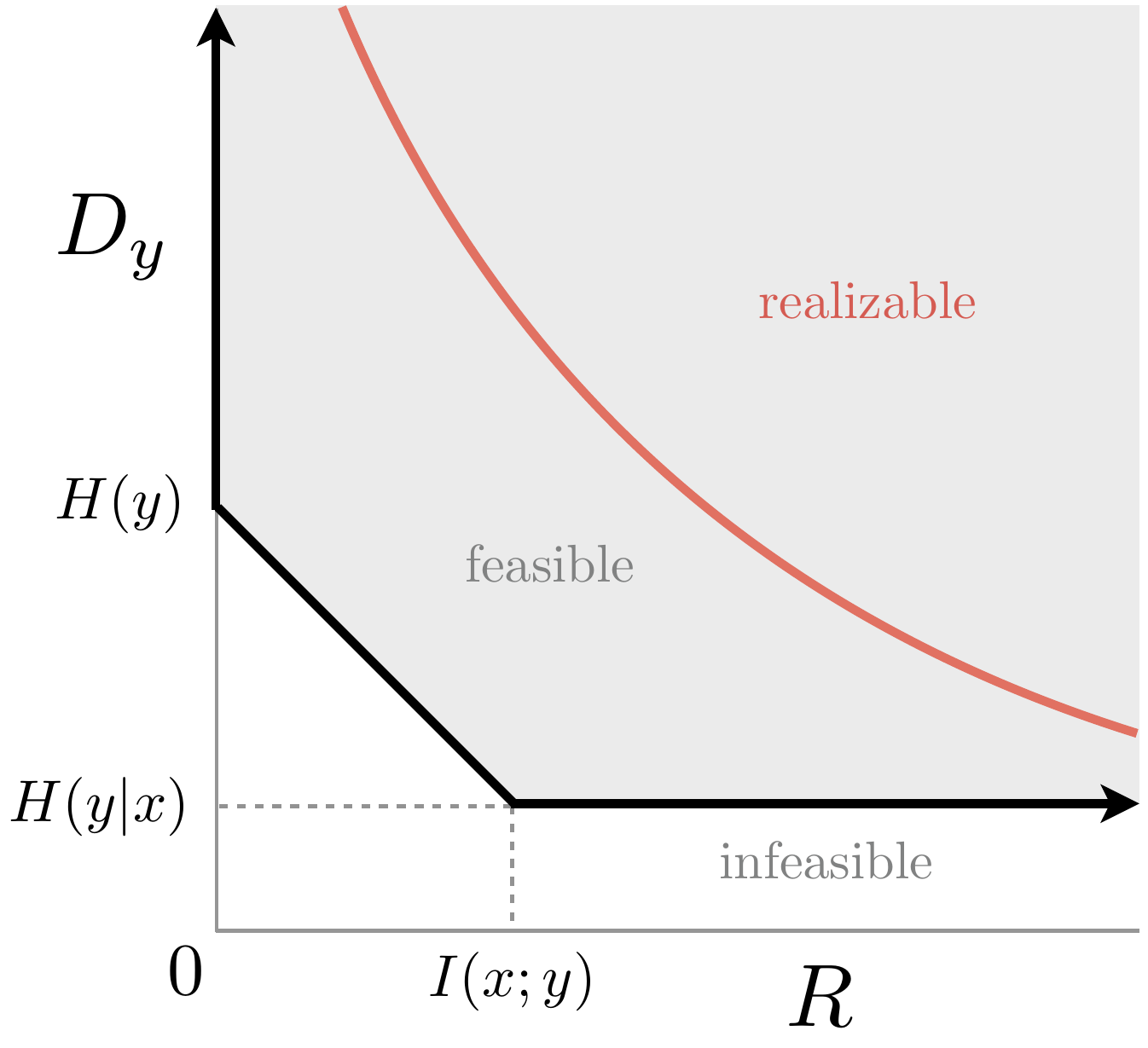}
   \caption{$R$-$D_y$ tradeoff for the supervised case.}
 \end{subfigure} \\[0.5cm]
 \begin{subfigure}[b]{.5\linewidth}
   \includegraphics[width=\textwidth]{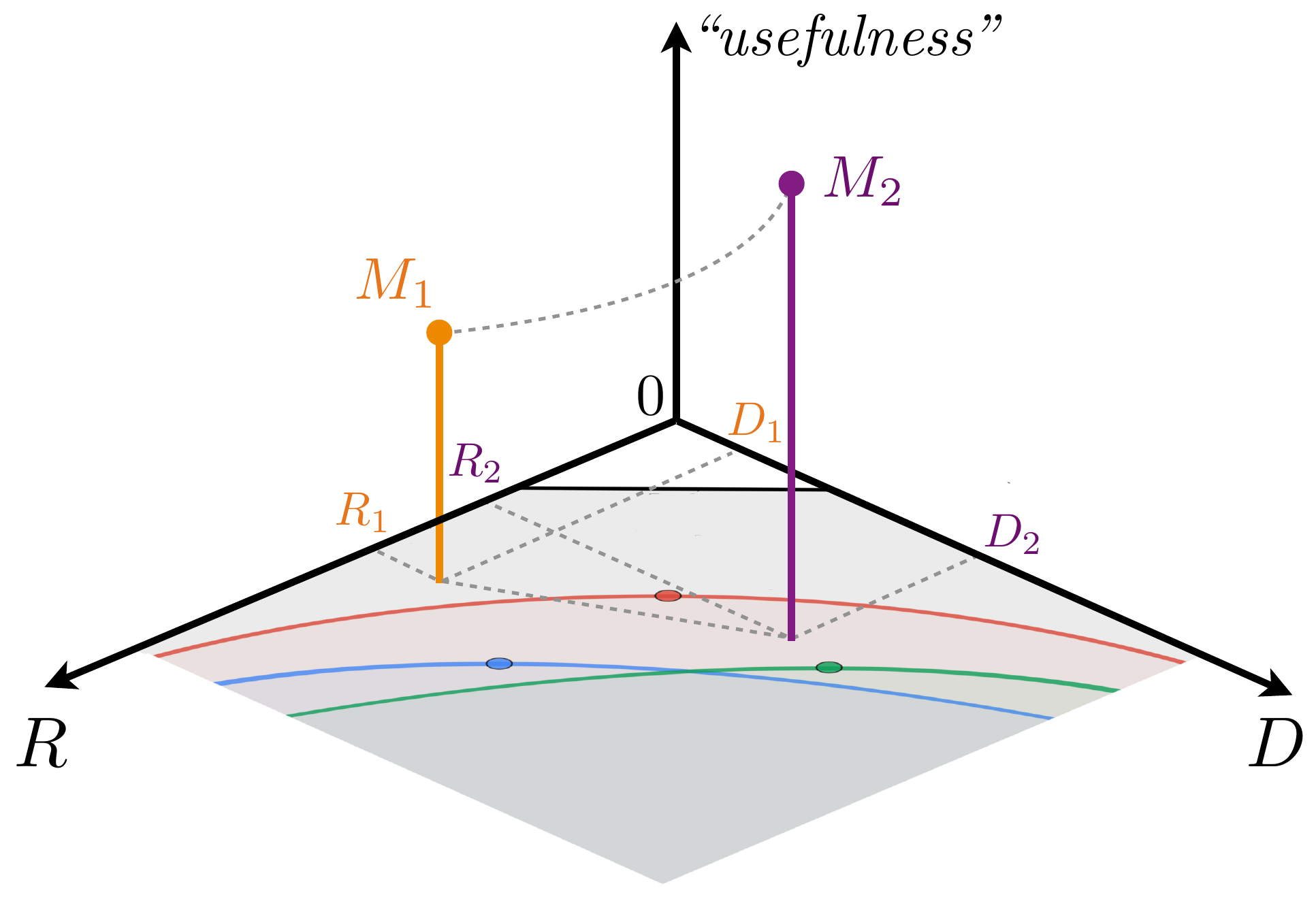}
   \caption{Rate-distortion-usefulness tradeoff.}
 \end{subfigure} 
 \caption{\label{fig:rd_curve} 
 Figure (a) shows the Rate-distortion ($R$-$D$) tradeoff from \cite{alemi2018fixing}, where $D$ corresponds to the reconstruction term in the ($\beta$-)VAE objective, and the rate to the KL term. Figure (b) shows a similar tradeoff for the supervised case considered in \cite{achille2018information, alemi2016deep}. The ELBO $-\Lvae=-(R+D)$ does not reflect the usefulness of the learned representation for an unknown downstream task (see text), as illustrated in Figure (c).}
\end{figure}

In this paper we provided an overview of existing work on autoencoder-based representation learning approaches. One common pattern 
is that methods targeting rather abstract meta-priors such as disentanglement (e.g., $\beta$-VAE \cite{higgins2016beta}) were only applied to synthetic data sets and very structured real data sets at low resolution.  In contrast, fully supervised methods, such as FaderNetworks~\cite{lample2017fader}, provide representations which capture subtle properties of the data, can be scaled to high-resolution data, and allow fine-grained control of the reconstructions by manipulating the representation. As such, there is a rather large disconnect between methods which have some knowledge of the downstream task and the methods which invent a proxy task based on a meta-prior.  In this section, we consider this aspect through the lens of rate-distortion tradeoffs based on appropriately defined notions of rate and distortion. Figure~\ref{fig:rd_curve} illustrates our arguments.

\textbf{Rate-distortion tradeoff for unsupervised learning.}
It can be shown that models based purely on optimizing the marginal likelihood might be completely useless for representation learning. We will closely follow the elegant exposition from~\citet{alemi2018fixing}. Consider the quantities 
\begin{alignat*}{3}
  H &= - \int p(x) \log p(x) \diff{x} 
  &&= \ex_{p(x)}[-\log p(x)] \\
  D &= - \iint p(x) q_\phi(z | x) \log p_\theta(x | z) \diff{x}\diff{z} 
  \quad&&= \ex_{p(x)}[\ex_{q_\phi(z | x)}[-\log p_\theta(x | z)]]\\
  R &= \iint p(x) q_\phi(z | x) \log \frac{q_\phi(z | x)}{p(z)}  \diff{x}  \diff{z}
  &&= \ex_{p(x)}[\dkl(q_\theta(z| x) \| p(z))]
\end{alignat*}
where $H$ corresponds to the \emph{entropy} of the underlying data source, $D$ the \emph{distortion} (i.e., the reconstruction negative $\log$-likelihood), and $R$ the \emph{rate}, namely the average relative KL divergence between the encoding distribution and the $p(z)$.
Note that the ELBO objective is now simply $\text{ELBO} = -\Lvae = -(D + R)$ (or $-(D + \beta R)$ for $\beta$-VAE). \citet{alemi2018fixing} show that the following inequality holds:
\[
  H - D \leq R.
\]

Figure~\ref{fig:rd_curve} shows the resulting rate-distortion curve from \citet{alemi2018fixing} in the limit of arbitrary powerful encoders and decoders. The horizontal line $(R, 0)$ corresponds to the setting where one is able to encode and decode the data with no distortion at a rate of $H$. The vertical line $(0, D)$ corresponds to the zero-rate setting and by choosing a sufficiently powerful decoder one can reach the distortion of $H$. 
A critical issue is that any point on the line $D = H - R$ achieves the same ELBO. As a result, models based purely on optimizing the marginal likelihood might be completely useless for representation learning~\cite{alemi2018fixing,frenecblog} as there is no incentive to choose a point with a high rate (corresponding to an informative code). This effect is prominent in many models employing powerful decoders which function close to the zero-rate regime (see Section~\ref{sec:structarch} for details). As a solution,~\citet{alemi2018fixing} suggest to optimize the same model under a constraint on the desired rate $\sigma$, namely to solve $\min_{\phi, \theta} D + |\sigma - R|$. However, is this really enough to learn representations useful for a specific downstream task?

\textbf{The rate-distortion-usefulness tradeoff.}
Here we argue that even if one is able to reach any desired rate-distortion tradeoff point, in particular targeting a representation with specific rate $R$, the learned representation might still be useless for a specific downstream task. This stems from the fact that 
\begin{enumerate}[label=(\roman*)]
\item it is unclear which part of the total information (entropy) is stored in $z$ and which part is stored in the decoder, and
 \item even if the information relevant for the downstream task is stored in $z$, there is no guarantee that it is stored in a form that can be exploited by the model used to solve the downstream task.
\end{enumerate}

For example, regarding (i), if the downstream task is an image classification task, the representation should store the object class or the most prominent object features. On the other hand, if the downstream task is to recognize relative ordering of objects, the locations have to be encoded instead. Concerning (ii), if we use a linear model on top of the representation as often done in practice, the representation needs to have structure amenable to linear prediction.

We argue that there is no \emph{natural} way to incorporate this desiderata directly into the classic $R$-$D$ tradeoff embodied by the ELBO. Indeed, the $R$-$D$ tradeoff per se does not account for \emph{what} information is stored in the representation and \emph{in what form}, but only for \emph{how much}.

Therefore, we suggest a third dimension, namely \emph{``usefulness''} of the representation, which is orthogonal to the $R$-$D$ plane as shown in Figure~\ref{fig:rd_curve}. Consider two models $M_1$ and $M_2$ whose rates satisfy $R_1 > R_2$ and $D_1 < D_2$ and which we want to use for the (a priori unknown) downstream task $y$ (say image classification). It can be seen that $M_2$ is more useful (as measured, for example, in terms of classification accuracy) for $y$ even though it has a smaller rate and and a larger distortion than $M_2$. This can occur, for example, if the representation of $M_1$ stores the object locations, but models the objects themselves with the decoder, whereas $M_1$ produces blurry reconstructions, but learns a representation that is more informative about object classes.

As discussed in Sections \ref{sec:regbased}, \ref{sec:structarch}, and \ref{sec:structprior}, regularizers and architecture design choices can be used to determine what information is captured by the representation and the decoder, and how it is modeled. Therefore, the regularizers and architecture not only allow us to navigate the $R$-$D$ plane but simultaneously also the ``usefulness'' dimension of our representation. As usefulness is always tied to (i) a task (in the previous example, if we consider localization instead of classification, $M_1$ would be more useful than $M_2$) and (ii) a model to solve the downstream task, this implies that one cannot \emph{guarantee} usefulness of a representation for a task unless it is known in advance. Further, the better the task is known the easier it is to come up with suitable regularizers and network architectures, the extreme case being the fully supervised one. On the other hand, if there is little information one can rely on a generic meta-prior that might be useful for many different tasks, but will likely not lead to a very good representation for all the tasks (recall that the label-based FaderNetwork \cite{lample2017fader} scales to higher-resolution data sets than $\beta$-VAE \cite{higgins2016beta} which is based on a weak disentanglement meta-prior). How well we can navigate the ``usefulness'' dimension in Figure~\ref{fig:rd_curve} (c) is thus strongly tied to the amount of prior information available.

\paragraph{A rate-distortion tradeoff for supervised learning.} 
For arbitrary downstream tasks it is clear that it is hard to formalize the ``usefulness'' dimension in Figure~\ref{fig:rd_curve}. 
However, if we consider a subset of possible downstream tasks, then it may be possible to come up with a formalization.
In particular, for the case where the downstream task is to reconstruct (predict) some auxiliary variable $y$, we formulate an $R$-$D$ tradeoff similar to the one of ~\citet{alemi2018fixing} for a fully supervised scenario involving labels, and show that in this case, the $R$-$D$ tradeoff naturally reflects the usefulness for the task at hand. Specifically, we rely on the variational formulation of the information bottleneck principle proposed by \cite{achille2018information, alemi2016deep}. Using the terminology of \cite{achille2018information}, the goal in supervised representation learning is to learn a \emph{minimal} (in terms of code length) representation $z$ of the data $x$ that is \emph{sufficient} for a task $y$ (in the sense that it contains enough information to predict $y$). This can be formulated using the \emph{information bottleneck (IB) objective} \cite{tishby2000information} $\max_z I(y;z) - \beta I(z;x)$, where $\beta > 0$. 
By introducing parametrized distributions $p_\theta(y|z)$, $q_\phi(z|x)$ as in the derivation of VAEs (see Section~\ref{sec:prelim}) and by defining distortion as 
\begin{equation*}
D_y = - \iint p(x, y) q_\phi(z | x) \log p_\theta(y | z) \diff{x}\diff{y}\diff{z} = \ex_{p(x,y)}[\ex_{q_\phi(z | x)}[-\log p_\theta(y | z)]],
\end{equation*}
where $p(x,y)$ is the (true) joint distribution of $x$ and $y$ and $p(z)$ is a fixed prior, one obtains a variational approximation of the IB objective as $-(D_y + \beta R)$ (see \cite{achille2018information, alemi2016deep} for details).

Figure~\ref{fig:rd_curve} (b) illustrates the $R$-$D_y$ tradeoff. The best we can hope for is that $z$ stores all information about $y$ contained in $x$, i.e., $R=I(x;y)$. In the limit of arbitrarily complex $p_\theta(x|z)$ such a $z$ yields the minimum achievable distortion, which corresponds to the conditional entropy of $y$ given $x$, $H(y|x)$. As the rate decreases below $I(x;y)$ the distortion inevitably increases. When $R=0$ the code does not store any information and we have $p_\theta(y | z)=p_\theta(y)$, and hence for arbitrarily complex $p_\theta(x|z)$, $D_y=\ex_{p(x,y)}[\ex_{q_\phi(z | x)}[-\log p_\theta(y | z)]] = \ex_{p(y)}[-\log p_\theta(y)]] = H(y)$. As in the rate-distortion tradeoff for VAEs, all these extreme points are only achievable in the limit of infinite capacity encoders $p_\theta(x|z)$ and decoders $q_\phi(z|x)$. In practice, only models with a larger optimal IB objective $-(D_y + \beta R)$ are realizable.

In the supervised case considered here, the distortion corresponds to the $\log$-likelihood of the target $y$ predicted from the learned representation $z$. Therefore, given a model trained for a specific point in the $R$-$D_y$ plane, we know the predictive performance in terms of the negative $\log$-likelihood (or, equivalently, the cross-entropy) of that specific model.

Finally, we note that the discussed rate-distortion tradeoffs for the 
unsupervised and supervised scenario can be unified into a single 
framework, as proposed by \citet{alemi2018therml}. The resulting 
formulation recovers models such as semi-supervised VAE besides ($\beta$-)VAE, VIB, and 
Information dropout, but is no longer easily accessible through a 
two-dimensional rate-distortion plane. \citet{alemi2018therml} further establish 
connections of their framework to the theory of thermodynamics.

\section{Conclusion and Discussion}
Learning useful representations with little or no supervision is a key challenge towards applying artificial intelligence to the vast amounts of unlabelled data collected in the world. We provide an in-depth review of recent advances in representation learning with a focus on autoencoder-based models. In this study we consider several properties, \emph{meta-priors}, believed useful for downstream tasks, such as disentanglement and hierarchical organization of features, and discuss the main research directions to enforce such properties. In particular, the approaches considered herein either (i) regularize the (approximate or aggregate) posterior distribution, (ii) factorize the encoding and decoding distribution, or (iii) introduce a structured prior distribution. Given the current landscape, there is a lot of fertile ground in the intersection of these methods, namely, combining regularization-based approaches while introducing a structured prior, possibly using a factorization for the encoding and decoding distributions with some particular structure. 

Unsupervised representation learning is an ill-defined problem if the downstream task can be arbitrary.
Hence, all current methods use strong inductive biases and modeling assumptions. Implicit or explicit supervision remains a key enabler and, depending on the mechanism for enforcing meta-priors, different degrees of supervision are required. One can observe a clear tradeoff between the degree of supervision and how useful the resulting representation is: On one end of the spectrum are methods targeting  abstract meta-priors such as disentanglement (e.g., $\beta$-VAE \cite{higgins2016beta}) that were applied mainly to toy-like data sets. On the other end of the spectrum are fully supervised methods (e.g., FaderNetworks \cite{lample2017fader}) where the learned representations capture subtle aspects of the data, allow for fine-grained control of the reconstructions by manipulating the representation, and are amenable to higher-dimensional data sets.
Furthermore, through the lens of rate-distortion we argue that, perhaps unsurprisingly, maximum likelihood optimization alone can't guarantee that the learned representation is useful at all. One way to sidestep this fundamental issue is to consider the "usefulness" dimension with respect to a given task (or a distribution of tasks) explicitly. 

\bibliographystyle{IEEEtranN}
\bibliography{main}

\newpage

\appendix
\section{Estimators for MMD and HSIC} \label{sec:estimators}
Expanding \eqref{eq:mmdexact} and estimating $\mu_{p_x}, \mu_{p_y}$ as means over samples $\{x^{(i)}\}_{i=1}^N, \{y^{(i)}\}_{i=1}^M$, one obtains an unbiased estimator of the MMD as
\begin{align} \label{eq:mmdest}
    \widehat{\mmd}(p_x,p_y) &= \frac{1}{N(N-1)} \sum_{i=1}^N\sum_{j\neq i}^N k\left(x^{(i)},x^{(j)}\right) +  \frac{1}{M(M-1)} \sum_{i=1}^M \sum_{j\neq i}^M k\left(y^{(i)},y^{(j)}\right) \\
    &\qquad -  \frac{2}{NM} \sum_{i=1}^N \sum_{j=1}^M k\left(x^{(i)},y^{(j)}\right). \nonumber
\end{align}

The Hilbert-Schmidt independence criterion (HSIC) is a kernel-based independence criterion with the same underlying principles as the MMD. Given distributions $p_x(x)$ and $p_y(y)$ the goal is to determine whether $p(x, y) = p_x(x)p_y(y)$ and measure the degree of dependence. 
Intuitively, if the distributions $p_x$ and $p_y$ are parametrized with parameters $\alpha$ and $\beta$, i.e. $p_x=p_\alpha$ and $p_y = p_\beta$ minimizing $\hsic(p_\alpha,p_\beta)$ w.r.t. $\alpha$ and $\beta$ encourages independence between $p_\alpha$ and $p_\beta$. 
Given samples $\{x^{(i)}\}_{i=1}^N, \{y^{(i)}\}_{i=1}^N$ from two distributions $p_x$ and $p_y$ on $\mc X$ and $\mc Y$, and kernels $k\colon \mc X \to \mc X$ and $\ell \colon \mc Y \to \mc Y$, the HSIC can be estimated as
\begin{align} \label{eq:hsicest}
  \widehat{\hsic}(p_x,p_y) &=
  \frac{1}{N^2} \sum_{i,j} k\left(x^{(i)},x^{(j)}\right) \ell\left(y^{(i)},y^{(j)}\right) + 
\frac{1}{N^4} \sum_{i,j,k,l}^N k\left(x^{(i)},x^{(j)}\right) \ell\left(y^{(k)},y^{(l)}\right) \nonumber \\
    & \qquad- \frac{2}{N^3} \sum_{i,j,k}^N k\left(x^{(i)},x^{(j)}\right) \ell\left(y^{(i)},y^{(k)}\right).
\end{align}
We refer to \citet[Section~2.2]{lopez2018information} for a detailed description and generalizations.

\newpage
\section{Overview table}
\begin{table}[h!]
\input{tables/bigtable}

\end{table}

\end{document}

%% file: tables/metaprior.tex
{\small\centering
\begin{tabular}{ p{3.7cm} p{9.4cm}}
\toprule
Meta-prior & Methods \\
  \midrule
  Disentanglement & $\beta$-VAE \eqref{eq:betavae} \cite{higgins2016beta}, FactorVAE \eqref{eq:factorvae} \cite{kim2018disentangling}, $\beta$-TCVAE \eqref{eq:tcvae} \cite{chen2018isolating}, InfoVAE \eqref{eq:lvaeinfovae} \cite{zhao2017infovae}, DIP-VAE \eqref{eq:dipvae} \cite{kumar2018variational}, HSIC-VAE \eqref{eq:hsicvae} \cite{lopez2018information}, HFVAE \eqref{eq:hfvae} \cite{esmaeili2018structured}, VIB \cite{alemi2016deep}, Information dropout \eqref{eq:infodrop} \cite{achille2018information}, DC-IGN \cite{kulkarni2015deep}, FaderNetworks \eqref{eq:fadernets} \cite{lample2017fader}, VFAE \eqref{eq:vfae} \cite{louizos2015variational} \\[0.2cm]
  Hierarchical representation\tablefootnote{While PixelGAN-AE \cite{makhzani2017pixelgan}, VLAE \cite{chen2016variational}, and VQ-VAE \cite{van2017neural} do not explicitly model a hierarchy of latents, they learn abstract representations capturing global structure of images \cite{makhzani2017pixelgan, chen2016variational} and speech signals \cite{van2017neural}, hence internally representing the data in a hierarchical fashion.} & PixelVAE \cite{gulrajani2016pixelvae}, LVAE \cite{sonderby2016ladder}, VLaAE \cite{zhao2017learning}, Semi-supervised VAE \cite{kingma2014semi}, PixelGAN-AE \cite{makhzani2017pixelgan}, VLAE \cite{chen2016variational}, VQ-VAE \cite{van2017neural}
  \\[0.2cm]
  Semi-supervised learning & Semi-supervised VAE \cite{kingma2014semi}, \cite{narayanaswamy2017learning}, PixelGAN-AE \eqref{eq:pixelgan} \cite{makhzani2017pixelgan}, AAE \eqref{eq:aae} \cite{makhzani2015adversarial} \\[0.2cm]
  Clustering & PixelGAN-AE \eqref{eq:pixelgan} \cite{makhzani2017pixelgan}, AAE \eqref{eq:aae} \cite{makhzani2015adversarial}, JointVAE \cite{dupont2018joint}, SVAE \cite{johnson2016composing} \\
  \bottomrule
\end{tabular}
}

%% file: tables/disentanglement_table.tex
\setlength{\tabcolsep}{1.5mm}
\begin{tabular}{ l l l l c}
\toprule
  \textsc{Work} & $\mc L_\cdot$ & $R_1$ & $R_2$ & \textsc{Y} \\
  \midrule
  $\beta$-VAE \cite{higgins2016beta} & VAE & $\dkl(q_\phi(z|x)\|p(z))$ & & \\
  VIB \cite{alemi2016deep} & VAE & $\dkl(q_\phi(z|x)\|p(z))$ & & O\\
  PixelGAN-AE\cite{makhzani2017pixelgan} & VAE & $-I_{q_\phi}(x;z)$ & & O \\
  InfoVAE \cite{zhao2017infovae} & VAE  & $\dkl(q_\phi(z|x)\|p(z))$ & $\dkl(q_\phi(z)\|p(z))$  & \\
  Info. dropout \cite{achille2018information} & VAE  & $\dkl(q_\phi(z|x)\|p(z))$ & $\tc(q_\phi(z))$ & O\\
  HFVAE \cite{esmaeili2018structured} & VAE  & $-I_{q_\phi}(x;z)$ & $R_{\mc G}(q_\phi(z))\! +\! \lambda_2' \sum_{G \in \mc G} R_{\mc G}(q_\phi(z))$ & \\
  FactorVAE \cite{kim2018disentangling,chen2018isolating} & VAE  & & $\tc(q_\phi(z))$ &\\
  DIP-VAE \cite{kumar2018variational} & VAE  & & $\|\cov_{q_\phi(z)}[z] - I\|_\text{F}^2$ & \\
  HSIC-VAE \cite{lopez2018information} & VAE  & & $\text{HSIC}(q_\phi(z_{G_1}),q_\phi(z_{G_2}))$ & O \\
  VFAE \cite{louizos2015variational} & VAE  & & $\mmd(q_\phi(z|s=0), q_\phi(z|s=1))$ & \checkmark \\
  DC-IGN \cite{kulkarni2015deep} & VAE  & & & \checkmark \\ \midrule
FaderNet. \cite{lample2017fader}; \cite{hadad2018two}\tablefootnote{\citet{lample2017fader,hadad2018two} do not enforce a prior on the latent distribution and therefore cannot generate unconditionally.} & AE & 
$-\ex_{\hat p(x,y)}[\log P_\psi(1-y | E_\phi(x))]$ 
& & \checkmark \\
AAE/WAE \cite{makhzani2015adversarial,tolstikhin2017wasserstein} & AE & & $D_\text{JS}(E_\phi(z)\|p(z))$ & O \\
  \bottomrule
\end{tabular}

%% file: figures/disentanglement_picture.tex
\centering
\tikzstyle{encdiststy}=[draw, fill=blue!20, text width=11em,
    minimum height=2em, rounded corners]
\tikzstyle{decdiststy}=[draw, fill=green!20, text width=11.5em,
     minimum height=2em, rounded corners]

\tikzstyle{encsty}=[draw, text width=11em,
     text centered, minimum height=2em, rounded corners] 
\tikzstyle{decsty}=[draw, text width=11.5em,
     text centered, minimum height=2em, rounded corners] 

\newcommand{\regsize}{\small}
\newcommand{\encstrucsize}{\small}
\newcommand{\encsize}{\large}

\tikzstyle{rencsty}=[draw, fill=red!20, 
     text centered, minimum height=2em, rounded corners]

\tikzstyle{reglinesty}=[dashed]

\tikzstyle{aggsty}=[draw, fill=white, 
     text centered, minimum height=2em, rounded corners]
\tikzstyle{raggsty}=[draw, fill=orange!20, 
     text centered, minimum height=2em, rounded corners]
     
\tikzstyle{raggstymom}=[draw, fill=yellow!20, 
     text centered, minimum height=2em, rounded corners]

\makebox[\textwidth][c]{
\begin{tikzpicture}[scale=0.95]
     \coordinate (cencdesc) at (-2.3,2.3);
     \coordinate (cencdist) at (2,2.25);
     \coordinate (cdecdist) at (7,2.25);
     \coordinate (cx) at (-1,0);
     \coordinate (cenc) at (2,0);
     \coordinate (cdec) at (7,0);
     \coordinate (cxhat) at (10,0);

     \coordinate (crenc1) at (0,-1.5);
     \coordinate (crenc2) at (4,-1.5);
     \coordinate (crenc3) at (7.7,-1.5);

     \coordinate (cragg1) at (-2,-3);
     \coordinate (cragg2) at (0.8,-3);
     \coordinate (cragg3) at (4,-3);
     \coordinate (cragg7) at (8.5,-3);
     \coordinate (cragg4) at (-0.3,-4.5);
     \coordinate (cragg5) at (3.6,-4.5);
     \coordinate (cragg6) at (8.1,-4.5);
     
     \coordinate (cleg) at (4.2,-6);

  \node[left] at (cx) {\encsize$x$};
  \node[encsty] (enc) at (cenc) {\encsize$q_\phi(z|x)$};
  \node[decsty] (dec) at (cdec) {\encsize$p_\theta(x|z)$};
  \node[right] at (cxhat) {\encsize$\hat x$};
  \draw[->] (cx) to (enc.west);
   \draw[->] (enc.east) to (dec.west);
   \draw[->] (dec.east) to (cxhat);

   \node[aggsty] (r2) at (crenc2) {\regsize$\exe[q_\phi(z|x)]$};

\node[raggstymom] (ra4) at (cragg4) 
{\regsize$\text{HSIC}(q_\phi(z_{G_1}),q_\phi(z_{G_2}))$};
   \node[raggstymom] (ra5) at (cragg5) {\regsize$\|\cov_{q_\phi(z)}[z] - 
I\|_\text{F}^2$};
   \node[raggstymom] (ra6) at (cragg6) {\regsize$\mmd(q_\phi(z|s=0), 
q_\phi(z|s=1))$};

\draw[dashed] (ra4.north) to (r2.south);
\draw[dashed] (ra5.north) to (r2.south);
\draw[dashed] (ra6.north) to (r2.south);

\node[raggsty] (ra1) at (cragg1) {\regsize$\tc(q_\phi(z))$};
   \node[raggsty] (ra2) at (cragg2) {\regsize$\dkl(q_\phi(z)\|p(z))$};
   \node[raggsty] (ra3) at (cragg3) {\regsize$D_\text{JS}(q(z)\|p(z))$};
\node[raggsty] (ra7) at (cragg7) {$R_{\mc G}(q_\phi(z))\! +\! \lambda_2' \sum_{G \in \mc G} R_{\mc G}(q_\phi(z))$};
   \draw[dashed] (ra1.north) to (r2.south);
\draw[dashed] (ra2.north) to (r2.south);
\draw[dashed] (ra3.north) to (r2.south);
\draw[dashed] (ra7.north) to (r2.south);

\node[rencsty] (r1) at (crenc1) {\regsize$\dkl(q_\phi(z|x)\|p(z))$};
   \node[rencsty] (r3) at (crenc3) {\regsize$-I_{q_\phi}(x;z)$};

\draw[dashed] (r1.north) to (enc.south);
\draw[dashed] (r2.north) to (enc.south);
\draw[dashed] (r3.north) to (enc.south);

\node (nleg) at (cleg) {\crule[orange!20]{0.25cm}{0.25cm} Divergence-based regularizers of $q_\phi(z)$ \qquad \crule[yellow!20]{0.25cm}{0.25cm} Moment-based regularizers of $q_\phi(z)$};

\end{tikzpicture}}

%% file: tables/structured_arch.tex
\setlength{\tabcolsep}{1.5mm}
\begin{tabular}{ l l l c c c c c c c}
\toprule
  \textsc{Work} & \textsc{Enc} & \textsc{Dec} & \textsc{$P(z)$} & \textsc{Y} \\
  \midrule
  LadderVAE \cite{sonderby2016ladder}  & H & H & $\N$ & \\
  Variational LadderVAE\cite{zhao2017learning} & H & H & $\N$ & \\
  PixelVAE \cite{gulrajani2016pixelvae}  & H & H+A & $\N$ & \\
  Semi-supervised VAE \cite{kingma2014semi}  & H & & $\N$+$\C$ & \checkmark \\
  VLAE \cite{chen2016variational}  & & A & $\N$/L & \\
  \bottomrule
\end{tabular}

%% file: tables/structured_prior.tex
\setlength{\tabcolsep}{1.5mm}
\begin{tabular}{ l l l c c c c c c c}
  \toprule
    \textsc{Work} & \textsc{Enc} & \textsc{Dec} & \textsc{$p(z)$} & \textsc{Y} \\
  \midrule
  SVAE \cite{johnson2016composing} & & & G/M & \\
  VQ-VAE \cite{van2017neural}  &  & (A) & $\C$/L & & \\
  \cite{narayanaswamy2017learning} & & & G & \checkmark \\
  JointVAE \cite{dupont2018joint}  & & & $\C$+$\N$ & & \\
  \bottomrule
\end{tabular}

%% file: tables/bigtable.tex
\centering
\caption{\label{tab:infothcomp} Summary of the most important models considered in this paper. The objective is given by $\mc L_\cdot (\theta, \phi) + \lambda_1 \exe[R_1(q_\phi(z|x))] + \lambda_2 R_2(q_\phi(z))$, where $q_\phi(z) = \exe[q_\phi(z|x)]$ is the aggregate posterior, $R_1$ and $R_2$ are regularizers, and $\lambda_1, \lambda_2 >0$ are the corresponding regularization weights. The detailed description of the regularizers is provided in Section~\ref{sec:regbased}. 
We indicate the structure of the encoding and decoding distribution as follows: (H) hierarchical, (A) autoregressive, (default) fully connected or convolutional feed-feed forward neural network). 
We indicate the prior distribution as follows: ($\N$) multivariate standard Normal, ($\C$) categorical, (M) mixture distribution, (G) graphical model, (L) learned prior. 
We indicate whether labels are used as follows: (\checkmark) Labels are required for (semi-)supervised learning, (O) labels can optionally be used for (semi-)supervised learning. }
\vspace{2mm}
\centering
\makebox[\textwidth][c]{
\setlength{\tabcolsep}{0.7mm}
\begin{tabular}{ l l l l c c c c c c c}
\toprule
\textsc{Work} & $\mc L_\cdot $ & $R_1$ & $R_2$ & 
\textsc{Enc} & 
\textsc{Dec} & 
\textsc{$p(z)$} & 
\textsc{Y} \\
  \midrule
\cite{higgins2016beta} & VAE & $\dkl(q_\phi(z|x)\|p(z))$ & & & & $\N$ & \\
\cite{alemi2016deep} & VAE & $\dkl(q_\phi(z|x)\|p(z))$ & & & & & O \\\cmidrule{4-8}
\cite{zhao2017infovae} & VAE & $\dkl(q_\phi(z|x)\|p(z))$ & $\dkl(q_\phi(z)\|p(z))$ & & & $\N$ & \\
\cite{achille2018information} & VAE & $\dkl(q_\phi(z|x)\|p(z))$ & $\tc(q_\phi(z))$ & & & & O \\ \cmidrule{2-8}
\cite{makhzani2017pixelgan} & VAE & $-I_{q_\phi}(x;z)$ & & & A & $\N$/$\C$ & O \\
\cite{esmaeili2018structured} & VAE & $-I_{q_\phi}(x;z)$ & $R_{\mc G}(q_\phi(z))\! +\! \lambda_2' \sum_{G \in \mc G} R_{\mc G}(q_\phi(z))$ & & & $\N$ & \\\cmidrule{2-8}
\cite{kim2018disentangling,chen2018isolating} & VAE & & $\tc(q_\phi(z))$ & & & $\N$ & \\
\cite{kumar2018variational} & VAE & & $\|\cov_{q_\phi(z)}[z] - I\|_\text{F}^2$ & & & $\N$ & \\
\cite{lopez2018information} & VAE & & $\text{HSIC}(q_\phi(z_{G_1}),q_\phi(z_{G_2}))$ & & & $\N$ & O \\
\cite{louizos2015variational} & VAE & & $\mmd(q_\phi(z|s=0), q_\phi(z|s=1))$ & & &  $\N$ & \checkmark \\\cmidrule{4-8}
\cite{kulkarni2015deep} & VAE & & & & & $\N$ & \checkmark \\
\cite{kingma2014semi} & VAE & & & H & & $\N$+$\C$ & \checkmark \\
\cite{chen2016variational} & VAE & & & & A & $\N$/L & \\
\cite{gulrajani2016pixelvae} & VAE & & & H & H+A & $\N$ & \\
\cite{sonderby2016ladder} & VAE & & & H & H & $\N$ & \\
\cite{zhao2017learning} & VAE & & & H & H & $\N$ & \\
\cite{johnson2016composing} & VAE & & & & & G/M & \\
\cite{narayanaswamy2017learning} & VAE & & & & & G & \checkmark \\
\cite{dupont2018joint} & VAE & & & & & $\C$+$\N$ & & \\
\cite{van2017neural} & VAE & & & & A & $\C$/L & & \\\midrule
\cite{lample2017fader,hadad2018two}\tablefootnote{\citet{lample2017fader,hadad2018two} do not enforce a prior on the latent distribution and therefore cannot generate unconditionally.} & AE & 
$-\ex_{\hat p(x,y)}[\log P_\psi(1-y | E_\phi(x))]$ 
& & & & & \checkmark \\
\cite{makhzani2015adversarial,tolstikhin2017wasserstein} & AE & & $D_\text{JS}(E_\phi(z)\|p(z))$ & & &  $\N$/$\C$/M & O \\
  \bottomrule
\end{tabular}}